\documentclass[nohyperref]{article}

\usepackage{microtype}
\usepackage{graphicx}
\usepackage{subfig}
\usepackage{booktabs} 

\usepackage{hyperref}

\usepackage[accepted]{icml2023}

\usepackage{amsmath}
\usepackage{amssymb}
\usepackage{mathtools}
\usepackage{amsthm}
\usepackage[utf8]{inputenc}
\usepackage{uclaml_macro}
\usepackage{tikz}
\usepackage{bbm}
\usepackage{marvosym}

\usepackage[textsize=tiny]{todonotes}

\usepackage[capitalize,noabbrev]{cleveref}

\newcommand{\ourmethod}{FedGMM}
\theoremstyle{plain}

\usepackage[textsize=tiny]{todonotes}
\usepackage{enumitem}

\icmltitlerunning{Personalized Federated Learning under Mixture of Distributions}

\begin{document}

\twocolumn[
\icmltitle{Personalized Federated Learning under Mixture of Distributions}



\icmlsetsymbol{equal}{*}

\begin{icmlauthorlist}
\icmlauthor{Yue Wu}{equal,yyy}
\icmlauthor{Shuaicheng Zhang}{equal,sch}
\icmlauthor{Wenchao Yu}{comp}
\icmlauthor{Yanchi Liu}{comp}
\icmlauthor{Quanquan Gu}{yyy}
\icmlauthor{Dawei Zhou}{sch}
\icmlauthor{Haifeng Chen}{comp}
\icmlauthor{Wei Cheng\textsuperscript{\Letter}}{comp}
\end{icmlauthorlist}

\icmlaffiliation{yyy}{Department of Computer Science, University of California, Los Angeles, USA. }
\icmlaffiliation{sch}{Department of Computer Science, Virginia Tech, Blacksburg, USA.}
\icmlaffiliation{comp}{NEC Laboratories America, Princeton, USA.}

\icmlcorrespondingauthor{Wei Cheng}{weicheng@nec-labs.com}

\icmlkeywords{Machine Learning, ICML}

\vskip 0.3in
]



\printAffiliationsAndNotice{\icmlEqualContribution} 

\begin{abstract}

The recent trend towards Personalized Federated Learning (PFL) has garnered significant attention as it allows for the training of models that are tailored to each client while maintaining data privacy. However, current PFL techniques primarily focus on modeling the conditional distribution heterogeneity (i.e. concept shift), which can result in suboptimal performance when the distribution of input data across clients diverges (i.e. covariate shift). Additionally, these techniques often lack the ability to adapt to unseen data, further limiting their effectiveness in real-world scenarios.
To address these limitations, we propose a novel approach, \ourmethod, which utilizes Gaussian mixture models (GMM) to effectively fit the input data distributions across diverse clients. The model parameters are estimated by maximum likelihood estimation utilizing a federated Expectation-Maximization algorithm, which is solved in closed form and does not assume gradient similarity. Furthermore, \ourmethod\ possesses an additional advantage of adapting to new clients with minimal overhead, and it also enables uncertainty quantification. Empirical evaluations on synthetic and benchmark datasets demonstrate the superior performance of our method in both PFL classification and novel sample detection.
\end{abstract}

\section{Introduction}

\label{sec:intro}

The sheer volume of data at our disposal today is often sequestered in isolated silos, making it challenging to access and utilize. Federated Learning (FL) presents a groundbreaking solution to this conundrum, enabling collaborative learning across distributed data sources without compromising the confidential nature of the original training data, while also being fully compliant with government regulations \cite{lim2020federated,aledhari2020federated,mothukuri2021survey}. This method has drawn a lot of attention in recent years since it enables model training on diverse, decentralized data while protecting privacy and security. In many applications, the model needs to be adjusted for each device or user, notably the \textit{cross-device} scenarios. These situations are the focus of Personalized Federated Learning (PFL), which tries to provide client-specific model parameters for a certain model architecture. In this scenario, each client aims to obtain a local model with a respectable test result on its own local data distribution~\cite{wang2019federated}.

In order to cater to the unique needs of individual clients and address the statistical diversity that exists among them, existing PFL studies frequently resort to an elegant amalgamation of federated learning and other sophisticated approaches, such as meta-learning~\cite{sim2019investigation}, client clustering~\cite{ghosh2020efficient}, multi-task learning~\cite{marfoq2021federated}, knowledge distillation~\cite{pmlr-v139-zhu21b}, and the lottery ticket hypothesis\cite{ChengICDM}, to achieve the desired level of personalization. For example, clients can be assigned to many clusters, and clients in the same cluster are assumed to use the same model via clustered FL techniques~\cite{ghosh2020efficient}. To train a global model as a meta-model and then fine-tune the parameters for each client, several researchers have embraced meta-learning based methodologies~\cite{sim2019investigation,jiang2019improving}. Wang et al.~\cite{ChengICDM} suggested utilizing a routing hypernetwork to expertly curate and assemble modular blocks from a globally shared modular pool, in order to craft bespoke local networks through the application of the lottery ticket theory. A recent study~\cite{marfoq2021federated} that leveraged the multi-task learning concept posited that each client's data distribution was a composite of $M$ underlying distributions, and proposed the use of a linear mixture model to make tailored decisions based on the shared components among them. It optimizes the varying conditional distribution $\PP_c(\yb|\xb)$ under the assumption that the marginal distributions $\PP_c(\xb) = \PP_{c'}(\xb)$ are the same for all clients (Assumption 2 in \cite{marfoq2021federated}). 

While these approaches are adept at addressing the issue of \textit{conditional distribution heterogeneity}, commonly referred to as \textit{concept shift}, within PFL, they fall short in addressing the more comprehensive issue of general statistical heterogeneity which encompasses other forms of variability, such as feature distribution skew (i.e., \textit{covariate shift})~\cite{kairouz2021advances}, that is each client has different input marginal distributions (i.e., $\PP_c(\xb) \neq \PP_{c'}(\xb)$). For example, even with handwriting recognition, users may exhibit variations in stroke length, slant, and other nuances when writing the same phrases. In reality, data on each client may be deviated from being identically distributed, say, $\PP_c \neq \PP_{c'}$ for clients $c$ and $c'$. That is, the joint distribution $\PP_c(\mathbf x,\mathbf y)$ (can be rewritten as $\PP_c(\mathbf y|\mathbf x)\PP_c(\mathbf x)$) may be different across clients. We refer to it as the ``\textit{joint distribution heterogeneity}" problem. Current approaches fall short of fully encapsulating the intricacies of the variations in the joint distribution among clients, owing to their tendency to impose a presumption of constancy on one term while adjusting the other~\cite{marfoq2021federated,pmlr-v139-zhu21b}.

Besides, cross-device federated learning applications are often faced with a phenomenon known as \textit{client drift}. This occurs when the learning model is deployed in a real-world online setting, and the distribution of inputs it encounters differs from the distribution it was trained on. As a result, the model's performance may be severely impacted. For instance, a PFL model trained on the historical medical records of a specific patient population may exhibit significant regional or demographic biases when tested on a new patient~\cite{shukla2019interpolation,purushotham2016variational}. To mitigate this, it is crucial to develop a cutting-edge PFL methodology that can easily adapt to new clients while incorporating the capability to perform uncertainty quantification. The key to achieving this lies in the ability to identify and account for any outliers that may deviate from the established training data distribution. Such a methodology would elevate PFL to a practical solution, enabling it to be deployed in a wide range of applications with confidence.

In this study, we propose a \underline{Fed}erated \underline{G}aussian \underline{M}ixture \underline{M}odel (\ourmethod) approach, which utilizes Gaussian mixture models to tackle the aforementioned issues. Our approach operates under the assumption that the joint distribution of data is a linear mixture of several base distributions. \ourmethod\ builds up PFL by maximizing the log-likelihood of the observed data. To maximize the log-likelihood of the mixture model, we suggest a federated Expectation-Maximization (EM) algorithm for model parameter learning. The update rule for the Gaussian components has a closed-form solution and does not resort to gradient methods. To ensure convergence of the EM update rule, we incorporate our algorithm with the theoretical analysis of federated EM for GMMs. The Gaussian parameters inferred by the server offer a detailed global statistical descriptor of the data, and can be applied for various purposes, including density estimation and clustering, etc.

To sum up, our contributions are as follows:\vspace{-0.1in}

\begin{itemize}[itemsep=1.5pt,topsep=0pt,parsep=0pt]


\item For the first time, this study explicitly addresses the challenging issue of joint distribution heterogeneity in PFL. Our approach serves as a novel solution to this problem, enabling the capability to perform uncertainty quantification. Furthermore, the proposed approach is designed to be highly flexible, allowing for easy inference of new clients, who did not participate in the training phase. This is achieved by learning their personalized mixture weights with a small computational overhead.

\item  Our method presents a highly adaptable framework that is independent of supervised discriminative learning models, making it easily adaptable to other learning models. The model parameters are learned in an end-to-end fashion via maximum likelihood estimation, specifically a federated Expectation-Maximization (EM) algorithm. Furthermore, we have theoretically analyzed the convergence bound of our log-likelihood function, providing a solid theoretical foundation for our approach. The federated learning process for the Gaussian mixture is a novel federated unsupervised learning approach, which may be of independent interest.

\item In the experiments, we assessed our technique on both artificial and real-world datasets to validate its efficacy in simulating the mixture joint distribution of PFL data for classification, as well as its capacity to discover novel samples. The outcomes show that our technique performs significantly better than the state-of-the-art (SOTA) baselines.

\end{itemize}

\vspace{-0.1in}
\section{Problem Formulation}
\vspace{-0.07in}
\label{sec:problem}
\paragraph{Notations} We use lowercase letters/words to denote scalars, lowercase bold letters/words to denote vectors, and uppercase bold letters to denote matrices. 
We use $\| \cdot \|$ to indicate the Euclidean norm. We also use the standard $O$ and $\Omega$ notations. 
For a positive integer $N$, $[N] := \{1,2,\dots,N\}$.


We focus on the personalized federated classification task. Suppose there exist $C$ clients. Each client $c \in [C]$ has its own dataset of size $N_c$, where a sample $\sbb_{c,i} = (\xb_{c,i}, \yb_{c,i})$ is assumed to be drawn from its distribution $\PP_c(\xb,\yb)$. The local data distribution $\PP_c(\xb,\yb)$ can be different. Therefore, it is natural to choose different hypotheses $h_c \in \cH$ for each client $c$. Here, $\cH$ can be some general and highly expressive function class like neural networks. 

In this work, we use $h_c(\xb,\yb)$ (sometimes denoted by $h_c(\sbb)$) to represent the likelihood of the sample $\sbb = (\xb,\yb)$. For classification tasks, the goal is naturally to achieve the expected maximum log-likelihood:
\vspace{-0.1in}
\begin{align*}
    \forall c \in [C], \qquad 
    \max_{h_c \in \cH}
    \mathop{\EE}_{(\xb,\yb) \sim \PP_c}
    \big[
    \log
    \big( 
    h_c(\xb,\yb)
    \big) 
    \big].
\end{align*}
\vspace{-0.2in}
\subsection{Mixture of Joint Distributions}
To facilitate federated learning, it is necessary to pose assumptions on how the distributions of different clients are similar, such that the data from one client can be utilized to improve the learning of other clients. To this end, we adopt the simple but general assumption that the distribution of one client is a mixture of several base distributions:
\begin{align}\label{eqn:jointGMM}
    \PP_c(\xb, \yb)
    & =
    \sum_{m=1}^{M} \pi^*_c(m) \PP^{(m)}(\xb, \yb), \forall c \in [C].
\end{align}
Here, $\PP^{(m)}$ denotes the $m$-th base distribution that is shared across all clients, while $\pi_c^*(m)$ can differ for different client $c$. With this presumption, we may benefit from the fact that any client can gain knowledge from datasets collected from all other clients but eschew clear statistical assumptions about local data distributions, and the heterogeneous joint distribution can be accurately modeled as well. This assumption in a federated setting was first introduced by \citet{marfoq2021federated} and was named FedEM. What differs is that \citet{marfoq2021federated} additionally assumes that the marginal distributions of each base distribution $\PP^{(m)}(\xb)$ are the same. This implies that every client has the same input distribution $\PP_c(\xb) = \PP_{c'}(\xb)$, while the conditional distributions $\PP_c(\yb|\xb)$ are different across different clients, and admit a form of linear mixtures.
\begin{align} \label{eqn:fedem-y-x}
    \PP_c(\yb|\xb) = \sum_{m=1}^{M} \pi^*_c(m) \PP^{(m)}(\yb|\xb).
\end{align}
This assumption simplifies what the clients must learn: the mixture weights $\pi^*_c(\cdot)$ and the conditional distribution $\PP^{(m)}(\yb|\xb)$. In other words, the training objective will degenerate to minimizing the cross entropy for classification, rather than to maximizing the likelihood of $\{(\xb_{c,i}, \yb_{c,i})\}_{i\in[N_c]}$. In contrast, if we allow $\PP^{(m)}(\xb)$ to be different, then the conditional probability will appear in the following form:
\begin{align} \label{eqn:fedgmm-y-x}
    \PP_c(\yb|\xb) = \frac{\sum_{m=1}^{M} \pi^*_c(m) \PP^{(m)}(\yb|\xb) \PP^{(m)}(\xb)}{\sum_{m=1}^{M} \pi^*_c(m) \PP^{(m)}(\xb)}.
\end{align}

\begin{figure}%
\vspace{-0.4cm}
    \centering
    \subfloat[\centering Same $\PP_c(\xb)$]{{\includegraphics[width=0.498\linewidth]{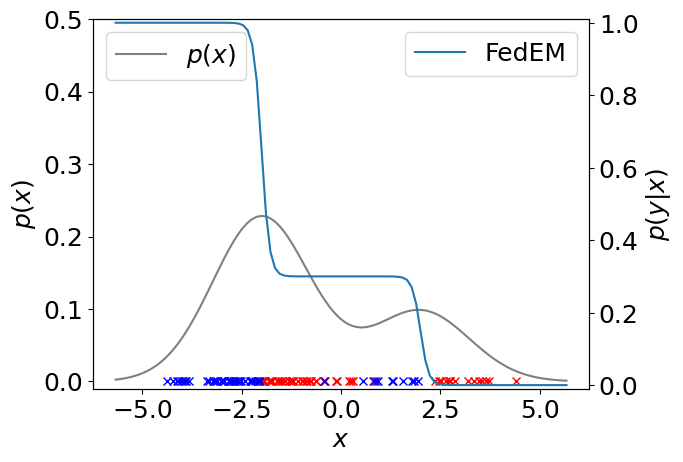} }}%
    \hspace{-2.5mm}
    \subfloat[\centering Different $\PP_c(\xb)$]{{\includegraphics[width=0.498\linewidth]{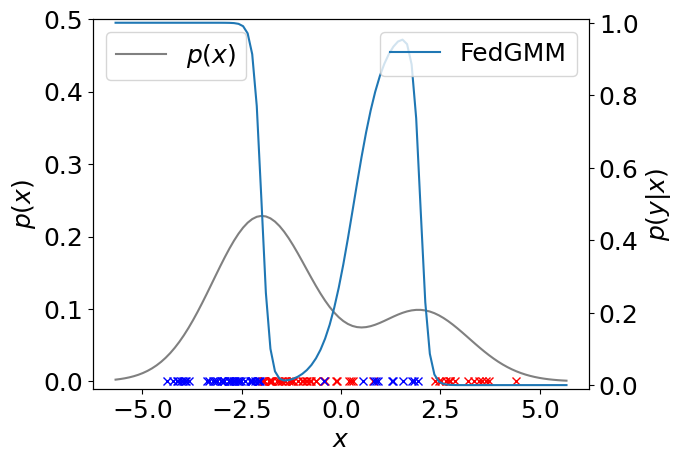} }}%
    \caption{An illustrative example: data are drawn from a mixture of two distributions: $\PP^{(1)}(x) = \cN(x;-2,1.5)$, $y = f^{(1)}(x) = \ind\{x<-2\}$ and $\PP^{(2)}(x) = \cN(x;2,1.5)$, $y = f^{(2)}(x) = \ind\{x<2\}$. Figure (a) shows how an algorithm that assumes $\PP^{(1)} = \PP^{(2)}$ fails to predict the label correctly. Figure (b) shows that once the input distribution is considered, the model can fully capture the data distribution.} 
    \label{fig:fedem-fedgmm}%
\end{figure}

It is clear that aside from learning the conditional distribution $\PP^{(m)}(\yb|\xb)$, to faithfully characterize the conditional probability, we also need to learn the base input distribution $\PP^{(m)}(\xb)$. Figure \ref{fig:fedem-fedgmm} shows that when $\PP^{(m)}(\xb)$ are indeed different, there will be a fundamental gap between the classification errors.

\section{Proposed Method}
\label{sec:method}
\subsection{Motivation}
It is widely known that the likelihood maximization problem under a linear mixture structure can be solved by the Expectation-Maximization (EM) technique. 
Consider the following learning objective: $\forall c \in [C]$,
\begin{align*}
    \max_{\pi_c, \btheta, \bphi}
    \mathop{\EE}_{(\xb, \yb) \sim \PP_c}
    \Big[
    \log
    \Big( 
    \sum_{m=1}^{M}
    \pi_c(m)
    P_{\bphi_{m}}(\xb)
    P_{\btheta_{m}}(\yb|\xb)
    \Big) 
    \Big].
\end{align*}
Similar to \citet{marfoq2021federated}, this kind of problem can be solved by optimizing the parameters $\bphi_m$ and $\btheta_m$ separately via gradient methods. The difficulty in learning $\PP^{(m)}(\xb)$ lies in that most modern density estimation models (such as auto-regressive models, normalizing flows, etc) are either very large, rendering it impractical for edge devices, or taking extremely long training time.

To learn the input distribution $\PP_c(x)$ efficiently, we resort to Gaussian mixture models (GMM); for the conditional distribution $\PP_c(y|x)$, we follow the same idea as \citet{marfoq2021federated}, to use light-weighted, parameterized supervised learning models.

\subsection{Models}
Formally, we define our model as:
\begin{itemize}[itemsep=2pt,topsep=0pt,parsep=0pt]
    \item All clients share the GMM parameters $\{ \bmu_{m_1}, \bSigma_{m_1} \}$ for any $m_1 \in [M_1]$. 
    \item All clients share the supervised learning parameters $ \btheta_{m_2}$ for $m_2 \in [M_2]$.
    \item Each client $c$ keeps its own personalized learner weights $\pi_c(m_1, m_2)$, which satisfies $\sum_{m_1, m_2}\pi_c(m_1, m_2) = 1$.
\end{itemize}

Note that $M_1$ is the number of Gaussian components, and $M_2$ is the number of learners. Under our definition of the models above, for client $c$, its hypothesis is defined as:
\begin{align*}
    h_c(\xb,\yb) 
    & := 
    \sum_{m_1, m_2}
    \pi_c(m_1, m_2)
    \cN(\xb; \bmu_{m_1}, \bSigma_{m_1})
    P_{\btheta_{m_2}}(\yb|\xb),
\end{align*}
where $\cN(\cdot; \bmu, \bSigma)$ denotes the probability density of multi-variate Gaussian distribution\footnote{ The probability density of multi-variate Gaussian is defined as:
$
    \cN(\xb; \bmu, \bSigma) 
    := 
    \frac{1}{\sqrt{(2\pi)^d \mathrm{det}(\bSigma)}}
    \exp
    \big(
    - \frac{1}{2}
    (\xb-\bmu)^{\top} \bSigma^{-1} (\xb-\bmu)
    \big).$
},
and $P_{\btheta}(\yb|\xb)$ is some supervised-learning model parameterized by $\btheta$.

Under this formulation, our optimization target becomes $\forall c \in \cC$ (we omit $M_1$ or $M_2$ when clear):
\scriptsize{
\begin{align*} 
    \max_{\pi_c, \btheta}
    \mathop{\EE}_{(\xb,\yb) \sim \PP_c}
    \Big[
    \log
    \Big( 
    \sum_{m_1, m_2}
    \pi_c(m_1, m_2)
    \cN(x; \bmu_{m_1}, \bSigma_{m_1})
    P_{\btheta_{m_2}}(\yb|\xb)
    \Big) 
    \Big].
\end{align*}
}
\vspace{-0.3in}
\normalsize
\subsection{The Centralized EM Algorithm}
To reduce notation clutter, we use $m = (m_1, m_2)$ and $\bTheta_{m} = ( \bmu_{m_1}, \bSigma_{m_1}, \btheta_{m_2} )$. We denote our model as $P_{\pi_c, \bTheta}(\xb,\yb) = \sum_{m} \pi_c(m) P_{\bTheta_{m}}(\xb,\yb)$. Under this simplified notation, we can derive the EM algorithm as follows.  Here we first provide a brief derivation of the centralized EM algorithm. Later on, we will extend it to the client-server EM algorithm in a federated setting.

Denote $q_{\sbb}(\cdot)$ as a probability distribution over $[M]$, where $\sbb = (\xb,\yb)$. Also, for each sample, we assume it is drawn by first sampling the latent random variable $z \sim \pi_c(\cdot)$ and then sampling $(\xb,\yb) \sim P_{\bTheta_z}(\xb,\yb)$. 

To derive the centralized EM algorithm, we can establish the following lower bound of the likelihood for a sample $(\xb,\yb)$:\vspace{-0.1in}
\scriptsize{
\begin{align}
    & \log
    \big( 
    P_{\pi_c, \bTheta}(\xb,\yb)
    \big) \notag \\
    & \ge
    \sum_{m \in [M]}
    q_{\sbb}(m)
    \log
    \bigg( 
    \frac{
    P_{\pi_c, \bTheta}(z=m,\xb,\yb)}
    {q_{\sbb}(m)}
    \bigg) 
    \notag \\
    & =
    \sum_{m \in [M]}
    q_{\sbb}(m)
    \bigg[
    \log
    \bigg( 
    \frac{
    P_{\pi_c, \bTheta}(z=m)}
    {q_{\sbb}(m)}
    \bigg) 
    +
    \log
    \big(P_{\pi_c, \bTheta}(\xb,\yb|z=m)
    \big)
    \bigg] 
    \notag \\
    & =
    \sum_{m \in [M]}
    q_{\sbb}(m)
    \bigg[
    \log
    \bigg( 
    \frac{
    \pi_c(m)}
    {q_{\sbb}(m)}
    \bigg) 
    +
    \log
    \big(P_{\bTheta_m}(\xb,\yb)
    \big)
    \bigg] 
    \label{eqn:1} \\
    & =
    \sum_{m \in [M]}
    q_{\sbb}(m)
    \bigg[
    \log
    \bigg( 
    \frac{
    P_{\pi_c, \bTheta}(z=m|\xb,\yb)}
    {q_{\sbb}(m)}
    \bigg) 
    +
    \log
    \big(\PP_{\pi_c, \bTheta}(\xb,\yb)
    \big)
    \bigg],\label{eqn:2}
\end{align}
}\vspace{-0.2in}
\normalsize

where the first inequality is due to Jensen's inequality. Equation~\eqref{eqn:1} comes from the first equation (the line directly above~\eqref{eqn:1}); Equation~\eqref{eqn:2} comes from the same line by decomposing $P_{\pi_c, \bTheta}(z=m)$ into the conditional probability.

The EM algorithm will try to maximize Equation \eqref{eqn:1} and ~\eqref{eqn:2} alternatively, to ensure the lower bound of the likelihood (also called evidence lower bound) is maximized. This leads to the following update form:
\begin{itemize}[itemsep=1pt,topsep=0pt,parsep=0pt]
    \item \textbf{E-Step:} Fix $\pi_c$ and $\bTheta$, maximize Equation~\eqref{eqn:2} via $q_{\sbb}(m)$ for each $\sbb = (\xb, \yb)$, we see the optimal solution will be \vspace{-0.1in}
    \begin{align*}
        q_{\sbb}(m) & = P_{\pi_c, \bTheta}(z=m|\xb, \yb)
        \\
        & \propto \PP_{\pi_c, \bTheta}(z=m,\xb, \yb)
        = \pi_c(m) \PP_{\bTheta_m}(\xb, \yb).
    \end{align*} \vspace{-0.1in}
    \item \textbf{M-Step:} Fix $q_{\sbb}(\cdot |\xb, \yb)$, maximize Equation~\eqref{eqn:1} via $\pi_c$ and $\bTheta$, we see the optimal solution will be \vspace{-0.1in}
    \begin{align*}
        \pi_c(m) & = \frac{1}{N_c} \sum_{i=1}^{N_c} q_{\sbb_i}(m),
        \\
        \bTheta_m & = 
        \arg \max_{\bTheta} \sum_{i=1}^{N_c} q_{\sbb_i}(m) \log( \PP_{\bTheta}(\xb_i, \yb_i)).
    \end{align*} \vspace{-0.2in}
\end{itemize}

Now we substitute $m = (m_1, m_2)$ and $\bTheta_{m} = \{ \bmu_{m_1}, \bSigma_{m_1}, \btheta_{m_2} \}$. We can index the base component $\PP_{m_1, m_2}(\xb, \yb) = \cN(\xb; \bmu_{m_1}, \bSigma_{m_1}) \cdot \PP_{\btheta_{m_2}}(\yb|\xb)$. 
Substituting the specific model into the EM update rules proposed before, we can write the update rule at step $t$ as:
\begin{itemize}[itemsep=1pt,topsep=0pt,parsep=0pt]
    \item \textbf{E-Step:} For each client $c \in [C]$, for each $i \in [N_c]$,\vspace{-0.1in}
    \small{
    \begin{align}
        q^{(t)}_{\sbb_{c,i}}(m_1, m_2) 
        & \propto \pi^{(t-1)}_c(m_1, m_2) \cN(\xb_{c,i}; \bmu^{(t-1)}_{m_1}, \bSigma^{(t-1)}_{m_1}) 
        \notag \\
        & \qquad \cdot \PP_{\btheta^{(t-1)}_{m_2}}(\yb_{c,i}|\xb_{c,i}). \label{eqn:e-step} \tag{E}
    \end{align}
    }\vspace{-0.1in}
    \normalsize
    \item \textbf{M-Step:}  For each client $c \in [C]$, $m_1 \in [M_1]$, $m_2 \in [M_2]$, \vspace{-0.1in}
    \small{
    \begin{align}
         & \pi^{(t)}_c(m_1, m_2)= \frac{1}{N_c} \sum_{i=1}^{N_c} q^{(t)}_{\sbb_{c,i}}(m_1, m_2) ,
        \label{eqn:m-step} \tag{M}\\
        & \bmu^{(t)}_{m_1,c}
        =
        \frac{\sum_{i=1}^{N_c} \sum_{m_2} q^{(t)}_{\sbb_{c,i}}(m_1, m_2) \xb_{c,i}}{\sum_{i=1}^{N_c} \sum_{m_2} q^{(t)}_{\sbb_{c,i}}(m_1, m_2)},
        \notag \\
        & \bSigma^{(t)}_{m_1}
        =
        \frac{\sum_{i=1}^{N_c} \sum_{m_2} q^{(t)}_{\sbb_{c,i}}(m_1, m_2) (\xb_{c,i} - \bmu^{(t)}_{m_1,c})(\xb_{c,i} - \bmu^{(t)}_{m_1,c})^{\top}}{\sum_{i=1}^{N_c}\sum_{m_2} q^{(t)}_{\sbb_{c,i}}(m_1, m_2)}, 
        \notag \\
        & \btheta^{(t)}_{m_2,c} = 
        \arg \max_{\btheta} \sum_{i=1}^{N_c} \sum_{m_1} q^{(t)}_{\sbb_{c,i}}(m_1, m_2) 
        \log( \PP_{\btheta}(\yb_i|\xb_i)). \notag 
    \end{align} 
    }\vspace{-0.1in}
    \normalsize
\end{itemize}
The update rule for $\bmu$ and $\bSigma$ in the M-step is obtained by explicitly solving the optimization problem.
Notice that for $\btheta_{m_2,c}$, the maximization objective is equivalent to the (weighted) cross-entropy loss for classification.

\subsection{The Client-Server EM Algorithm}
Federated learning restricts that each client can only access their own data. In this section, we describe how to extend the centralized EM algorithm to the federated client-server setting.  Equation~\eqref{eqn:e-step} and~\eqref{eqn:m-step} describes how the client should maintain their personalized weights $\pi^{(t)}_c$, their own estimation of the shared GMM bases $(\bmu^{(t)}_{m_1,c}, \bSigma^{(t)}_{m_1,c})$  and the base learners $\btheta^{(t)}_{m_2,c}$.
When a central server is present, each client shall send their own parameters to the server and the server will aggregate the parameters and broadcast the aggregated parameter back to all clients. The detailed federated algorithm \ref{alg:FedGMM} is included in Appendix \ref{appendix:alg}.
 
More specifically, at each round, (1) the central server broadcasts the aggregated base models to all clients; (2) each client locally updates the parameter of the base models and the mixture weights according to Equation~\eqref{eqn:e-step} and~\eqref{eqn:m-step}; (3) the clients send the updated components $(\bmu^{(t)}_{m_1,c}, \bSigma^{(t)}_{m_1,c}), \btheta^{(t)}_{m_2,c}$ and the summed response $\gamma^{(t)}_{c}(m_1, m_2) = \sum_{i \in [N_c]} q^{(t)}_{\sbb_{c,i}}(m_1, m_2)$ back to the server; 4) the server aggregates the updates as follows:\vspace{-0.05in}
\begin{align*}
    \bmu^{(t)}_{m_1}
    & =
    \frac{\sum_{c \in [C]} \sum_{m_2 \in [M_2] } \gamma^{(t)}_{c}(m_1, m_2) \bmu^{(t)}_{m_1,c} }{\sum_{c \in [C]}\sum_{m_2 \in [M_2] } \gamma^{(t)}_{c}(m_1, m_2) } ,
    \notag \\
    \bSigma^{(t)}_{m_1}
    & =
    \frac{\sum_{c \in [C]} \sum_{m_2 \in [M_2] } \gamma^{(t)}_{c}(m_1, m_2) \bSigma^{(t)}_{m_1,c} }{\sum_{c \in [C]} \sum_{m_2 \in [M_2] }  \gamma^{(t)}_{c}(m_1, m_2)} ,
  \notag \\
    \btheta^{(t)}_{m_2}
    & =
    \frac{\sum_{c \in [C]} \sum_{m_1 \in [M_1] } \gamma^{(t)}_{c}(m_1, m_2) \btheta^{(t)}_{m_2,c} }{\sum_{c \in [C]} \sum_{m_1 \in [M_1] } \gamma^{(t)}_{c}(m_1, m_2)} . \label{eqn:aggregation} 
\end{align*} \vspace{-0.1in}

\vspace{-0.3cm}
\subsection{Theoretical Guarantees}

Since most federated learning algorithms are gradient-based, their convergence analyses usually assume the gradients of different clients are similar. For small steps of updates, the averaged updated parameters can still enjoy a decrease in the training loss. This is not the case for our GMM updates, because the M-step uses the closed-form solution for each client and then aggregates them, which means the widely-adopted gradient-similarity assumption will not help.

What we present in the following is an analysis of purely federated Gaussian Mixture Models. 
The convergence guarantee for the gradient-updated parameter $\btheta$ will have identical assumptions and proof as in \citet{marfoq2021federated}. We choose to omit the convergence result for $\btheta$. 
When leaving $\btheta$ out, we obtain a pure unsupervised likelihood maximization algorithm~\ref{alg:FedGMM-GMM} in Appendix~\ref{appendix:alg}. The centralized version of it is exactly the classical EM algorithm for GMM. The federated learning process for the Gaussian mixture is a novel federated unsupervised learning approach, which may be of independent interest.

To show the convergence of the proposed client-server EM algorithm, we consider the case that $\bSigma_m$ is fixed to $\Ib$, and only $\bmu$ is updated and aggregated. This assumption is widely adopted in previous works regarding the convergence of EM algorithms for GMM. It is also well known that if the covariance matrix $\bSigma_m$ is not restricted, GMM can assign one component $\cN(\cdot; \bmu_m, \bSigma_m)$ to one single data point $\xb$ such that $\bmu_m = \xb$ and $\bSigma_m \rightarrow \mathbf{0}$, so that the likelihood goes to positive infinity. Assuming $\bSigma_m=\Ib$ prevents this kind of unwanted divergence.

\begin{theorem} \label{thm:gmm}
Denote $F(\mu_{1:M}, \pi_{1:C})$ as the log-likelihood function, then we have\vspace{-0.1in}
\begin{align*}
    \frac{1}{T}
    \sum_{t=1}^{T}
    |F(\mu^{(t)}_{1:M}, \pi^{(t)}_{1:C})
    -
    F(\mu^{(t-1)}_{1:M}, \pi^{(t-1)}_{1:C})|
    & = O(T^{-1}).
\end{align*}
\end{theorem}\vspace{-0.1in}
Theorem~\ref{thm:gmm} implies that the log-likelihood will finally converge to a maximum. The idea of the proof (details included in Appendix \ref{sec:appproof}) relies on the use of first-order surrogates of $F$ to establish that each M-step will always increase the log-likelihood.

\vspace{-0.1in}
\section{Experiments}
\label{sec:exp}
\subsection{Datasets}
\label{sec:dataset}
\textbf{Synthetic dataset.}
The synthetic dataset can be seen as a $d$-dimensional extension of Figure~\ref{fig:fedem-fedgmm}. More specifically, assume there are $M$ Gaussian components $\PP^{(m)}(\xb) = \cN(\xb;\bmu_m, \Ib_{d})$, with a corresponding labeling function $F^{(m)}(\xb) = \ind\{ (\xb - \bmu_m)^{\top} \vb_m > 0\}$, where $\bmu_m$ and $\vb_m$ are specified beforehand. For each client $c$, the data generation is as follows: 1). sample $\pi_c$ from the Dirichlet distribution $\mathrm{Dir}(\alpha)$ with $\alpha=0.4$ to serve as the heterogeneous mixture weight; 2). for each sample $i \in [N_c]$, first generate $z_i \sim \pi_c(\cdot)$; 3). then draw $\xb_i \sim \PP^{(z_i)}(\xb_i) = \cN(\xb;\bmu_{z_i}, \Ib_{d})$ and $y_i = F^{(z_i)}(\xb_i)$. For the experiments, we set $M=3$ and $d=32$. We generate $C=300$ clients and each client has around $N_c = 3000$ samples. We also compare on non-Gaussian synthetic data as shown in Appendix 
\ref{app:nonGaussian}.

\textbf{Real datasets.}
We also use three federated benchmark datasets spanning different machine learning tasks to evaluate the proposed approach: image classification on CIFAR-10 and CIFAR-100~\cite{krizhevsky2009learning}, handwriting character recognition on FEMNIST \cite{caldas2018leaf}. We preprocessed all the datasets in the same manner as previously in~\cite{marfoq2021federated} to build the testbed. To simulate the joint distribution heterogeneity, we sample 50\% of image data (denoted as $\mathcal{D}_2$, $\mathcal{D}= \mathcal{D}_1\cup \mathcal{D}_2$) to perform a two-step approach for prepossessing image data: 1) we simulate heterogeneity of $\PP_c(\xb)$ by transforming sampled images with 90-degree rotation, horizontal flip and inverse~\cite{journalsSurvey} (denoted as $T(\cdot)$); 2) we introduce heterogeneity in $\PP_c(\yb | \xb)$ by applying a randomly generated permutation (denoted as $P_A$) to the labels of the transformed image data.
Formally, the new dataset, denoted as $\hat{\mathcal{D}}$, is defined as follows: $\hat{\mathcal{D}}=\mathcal{D}_1\cup \{(T(\mathbf{x}), P_A(\mathbf{y}))| (\mathbf{x},\mathbf{y})\in \mathcal{D}_2\}.$
In this way, we can obtain data from different joint distributions. We create the federated setting of CIFAR-10 by distributing samples with the same label across
the clients according to a symmetric Dirichlet distribution with parameter 0.4, as in~\cite{marfoq2021federated}. CIFAR-100 data are distributed following~\cite{marfoq2021federated}. For all tasks, we randomly split
each local dataset into training (60\%), validation (20\%), and test (20\%) sets. 
In Table~\ref{tab:dataModel}, we summarize the
datasets, tasks, number of clients, the total number of samples, and backbone discriminative architectures. 
\begin{table*}[ht]\vspace{-0.2in}
\centering
  \caption{Datasets and models.}\label{tab:dataModel}
  \small{
  \begin{center}
\scalebox{0.8}{
  \begin{tabular}{|c| c| c| c |c |}
  \hline
\textbf{Dataset} & \textbf{Task} & \textbf{Number of clients} & \textbf{Number of samples} & \textbf{Backbone Supervised Model} \\\hline
Synthetic& Binary Classification & 300 & $\sim1,000,000$ & Linear sigmoid function \\\hline
CIFAR-10 & Image classification& 80 & 60,000 & MobileNet-v2\\\hline
CIFAR-100 & Image classification & 100 & 60,000 & MobileNet-v2\\\hline
FEMNIST& Handwritten character recognition & 539& 120,772&2-layer CNN + 2-layer FFN\\\hline
\end{tabular}
}
\end{center}
}\normalsize
\vspace{-0.6cm}
\end{table*}

\subsection{Baseline Methods}
To demonstrate the efficiency of our method, we compare the proposed FedGMM with the following baselines:
\begin{itemize}[itemsep=1pt,topsep=0pt,parsep=0pt]
\item Local: a personalized model trained only
on the local dataset at each client;
\item FedAvg~\cite{mcmahan2017communication}: a generic FL method that trains a unique global model for all clients;
\item FedProx~\cite{li2020federated}: a re-parametrization of FedAvg to tackle statistical heterogeneity in FL;
\item FedAvg+~\cite{jiang2019improving}: a modification of FedAvg with two stages of training and local tuning;
\item Clustered FL~\cite{sattler2020clustered}: a framework exploiting geometric properties of the FL loss surface which groups the client population into clusters using conditional distributions;
\item pFedMe~\cite{t2020personalized}: a  bi-level optimization PFL that decouples the optimization of personalized models from learning the global model;
\item FedEM~\cite{marfoq2021federated}: a federated multi-task learning approach assuming that
local data distributions are mixtures of underlying distributions. 
\end{itemize}

\vspace{-0.1in}
\subsection{Implementation Details}\vspace{-0.06in}
To properly initialize each base component of the GMM, we employ a Resnet18~\cite{he2016deep} encoder that has been pre-trained on the ImageNet dataset to encode input images and generate embeddings of dimension 512. Recognizing that high dimensionality can lead to increased computational complexity and reduced effectiveness of GMM, 
we utilize PCA~\cite{Jolliffe86} to project the encoded embeddings into a lower-dimensional space of 48. For the sake of fairness in comparison, it is important to note that the Resnet18 encoder and PCA are exclusively employed for preprocessing inputs of the GMM component, while the inputs for the supervised backbone are raw images.

For each method, we follow~\cite{marfoq2021federated} to tune the learning rate via grid search. In our experiments, the number of local epochs of each method is set to 1, the total communication round is set to 200, and the batch size is set to 128, as in~\cite{marfoq2021federated}. For a fair comparison, we adopt the same supervised backbone architecture for all baselines. More implementation details are included in Appendix \ref{sec:appImp}. We also 
analyze parameter sensitivity of \ourmethod\ in Appendix \ref{sec:appPrameter}.

\vspace{-0.1in}
\subsection{Classification}
\vspace{-0.05in}
\label{sec:classification}

\begin{table*}[ht]\vspace{-0.2in}
\centering
  \caption{Average test accuracy (\%) across clients.}\label{tab:result2}
  \begin{center}
\scalebox{0.85}{
  \small{
  \begin{tabular}{|c| c| c| c |c |c |c |c ||c|}
  \hline
Dataset & Local & FedAvg & FedProx & FedAvg+ & ClusteredFL & pFedMe & FedEM & \ourmethod(Ours)\\\hline
Synthetic & 57.52 & 53.21 & 52.70 & 53.41 & 53.12 & 53.91 & 65.61  &\textbf{72.02}
\\\hline
CIFAR10 & 19.96& 45.53& 37.0&34.33&  38.81& 23.51 & 49.12 & \textbf{52.96}\\\hline
CIFAR100 &  13.36 & 17.71  &  7.95 & 11.51 & 12.46 &  9.92 &   17.28          & \textbf{22.33} \\\hline
FEMNIST&62.39& 75.08& 32.84&57.99& 75.04&39.45&75.56&\textbf{79.49}\\\hline
\end{tabular}
}
}
\normalsize
\end{center}


\vspace{-0.6cm}
\end{table*}
The results are shown in Table~\ref{tab:result2}. The evolution of average test accuracy
over time for each experiment is shown in the Appendix. From the table, we observe that FedAvg surpasses Local, which indicates that federated training improves performance because of taking advantage of knowledge from other clients. However, personalized methods such as FedAvg+, ClusteredFL, and pFedMe perform worse than FedAvg because they only locally adjust the global model on each client. This strategy is not sufficient to capture the diversity of the joint distribution and cannot handle sample-specific personalization when samples come from different marginal distributions have varying labeling functions. ClusteredFL also fails to outperform FedAvg on all datasets, highlighting the importance of knowledge sharing between clusters for training good personalized models. FedEM, on the other hand, performs better than other PFL baselines on most datasets by effectively modeling the heterogeneity of conditional distributions. 
As shown in the table, \ourmethod\ outperforms all baselines, achieving 26.1\% and 9.8\% improvement on CIFAR-100 and Synthetic dataset respectively compared to the leading baselines. This is a result of its ability to construct personalized models based on the joint data distribution, effectively capturing the heterogeneity of each sample across different clients.

\vspace{-0.1in}
\subsection{Novel Sample Detection}
\vspace{-0.06in}

\begin{figure}[t]
\vspace{-0.1in}
 \center{\includegraphics[width=0.93\linewidth]  {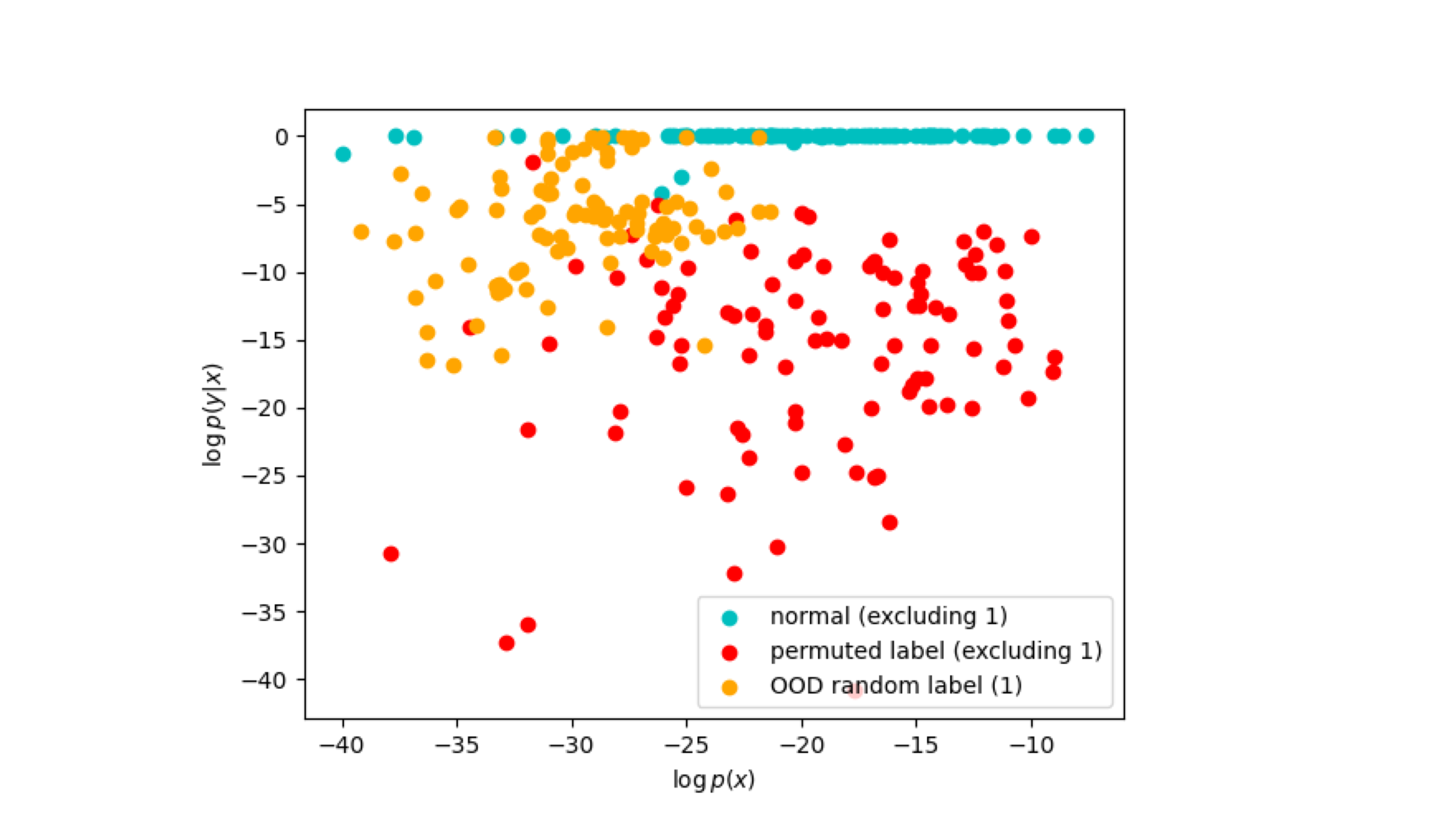}} \vspace{-0.2in}
 \caption{\label{fig:OOD} \ourmethod\ to detect marginal distribution and conditional distribution outliers. }\vspace{-0.5cm}
 \end{figure}

In our algorithm, the server meticulously maintains comprehensive, global statistics of all data points within the federated learning ecosystem, such as the GMM parameters\footnote{We can aggregate the global parameter $\pi  =  \sum_{c \in [C]} \frac{1}{N_c}\gamma_{c}$.} and the supervised learning components. Thus, for a new sample, the learned model is able to quickly infer its marginal distribution \footnote{The marginal distribution can be calculated by $\PP(\xb)=\sum_{m_2 \in [M2]}\sum_{m_1 \in [M1]}\pi(m_1,m_2) \cdot \cN \big(\xb; \bmu_{m_1}, \bSigma_{m_1} \big)$.}, conditional distribution (Eq. \ref{eqn:fedgmm-y-x}) and the joint distribution (Eq. \ref{eqn:jointGMM}). As such, a by-product of the model is that it can be used to detect out-of-distribution samples. We begin by using a typical leave-one-out method for out-of-distribution detection to demonstrate the effectiveness of our model in identifying various types of outliers. Specifically, we train our model using the MNIST dataset, with 50 clients each contributing 500 sampled images. In the training, we exclude images of number 1 and test on normal samples together with two types of outliers. The first category of outliers consists of images from the same marginal distribution $\PP(\xb)$, namely \{0, 2, 3,..., 9\}, but their labels have been altered by applying a random permutation. The second category of outliers are images of digit 1 that are not present in the training data.
We plot all the sample points with respect to their $\log\PP(\xb)$ and $\log\PP(\yb|\xb)$ values inferred by our model in Figure \ref{fig:OOD}. Here, the dots in cyan color are the normal ones. The orange points denote unseen input `1', and the red dots are outliers with the same marginal distribution but altered labels. We can observe that by modeling the conditional probability, the y-axis can separate red dots from the normal ones. Our density estimation model can separate the second type of outlier from other numbers as well.

To evaluate the performance of our OOD detection approach quantitatively, we trained each model using the following settings: we construct a federated setting using MNIST data, similar to the one described in Sec. \ref{sec:dataset}. Details are included in Appendix \ref{app:ood}. Basically, we create two sets of test samples drawn from the training distribution. The first set (as in-domain) remains unchanged. As the second (out-of-domain) set, we simulate the heterogeneity of $\PP_c(\xb)$ by transforming sampled images with a scale factor of 0.5, 90-degree rotation, and horizontal flip~\cite{journalsSurvey}. With the test samples, we want to investigate if a model can distinguish between known and novel samples.

For comparison purposes, since none of the baselines are able to detect novel samples, we adapt them as follows. Similar to the idea in ~\cite{liu2020energy}
, we use the prediction output logits with softmax to represent the classifier's confidence in different categories. The highest value among different categories is treated as the in-domain likelihood. This means the sharper the sample's prediction distribution, the more certain the classifier is that the sample is in-domain. Since the personalized baseline approaches do not have a global model, we selected the highest confidence value among different clients for a given new sample. It's worth noting that we did not include the Bayesian method in ~\cite{kotelevskii2022fedpop} as the baseline because the method can only perform novel detection at the client level, whereas here, we are conducting it at the sample level. Following \citep{cheng2021learning,vaze2022open,sharma2021identifying}, we report Area Under ROC (AUROC), Average Precision (AP), and Max-F1 for evaluation.

Table \ref{tab:oodauc} summarizes the results. We observe that \ourmethod\ outperforms all baseline's overall evaluation metrics, indicating the superiority of our model in modeling joint distribution. Our approach models each sample with a mixture distribution of different components, as described in Sec. \ref{sec:method}, which fits the mixture data well hence allowing to detect novel samples that are close to the boundary. Similar to \citep{liu2020energy}, in Figure \ref{fig:oodfre} in Appendix \ref{app:ood}, we visualize the normalized likelihood histogram of known and novel samples for \ourmethod, FedEM, and FedAvg. The figures indicate the likelihoods of \ourmethod\ are more distinguishable for known and novel samples than for the baselines.

\begin{table}[t]\vspace{-0.2cm}
\centering
  \caption{Comparison between \ourmethod\ and the applicable baselines on novel sample/client detection.}\label{tab:oodauc}
  \begin{center}
\scalebox{0.8}{
  \begin{tabular}{|c| c| c| c |}
  \hline
\textbf{Model} & \textbf{AUROC} & \textbf{AP} & \textbf{Max-F1} \\\hline
Local &50.74 &60.14 &66.67 \\\hline
FedAvg &66.55 & 68.05&66.67 \\\hline
FedProx &75.23 & 76.24&71.90 \\\hline
FedAvg+ & 66.65& 68.09& 66.67\\\hline
ClusteredFL & 50.74 &60.14 & 66.67 \\\hline
pFedMe & 73.32& 77.91&68.30 \\\hline
FedEM & 86.04 & 90.02 & 80.25 \\\hline
FedGMM & \textbf{99.21}& \textbf{99.60} & \textbf{99.49} \\\hline
\end{tabular}
}
\end{center}
\vspace{-0.6cm}
\end{table}


\subsection{Generalization to Unseen Clients}
As previously discussed, \ourmethod\ is flexible, enabling easy inference of new clients who did not participate in the training phase. This is accomplished by learning their personalized mixture weights. Specifically, we only need to update $q, \pi$, and $\gamma$ in lines 6, 8, and 10 of Algorithm \ref{alg:FedGMM} in Appendix \ref{appendix:alg}. All other parameters remain fixed during the update process. This adaptation incurs minimal computational costs.

To validate the effectiveness of our approach for generalization to unseen client data, we use the data with the same training setting as in the previous classification task (refer to Sec. \ref{sec:classification}). 
We use 80\% of clients to train the model and 20\% to test for unseen data adaptation, as per the setting in~\cite{marfoq2021federated}. We split samples into 50\% for adaptation and 50\% testing and adapt the mixture weights in our approach and the mixture weights of conditional distributions in FedEM using the adaptation samples from unseen clients. Aside from FedAvg+ and FedEM, it is uncertain how the other PFL algorithms can be adapted to unseen client. As FedAvg has a global model, we can still use it for test on the new data. As shown in Table. \ref{tab:unseen}, our approach obtains minimal decrease in accuracy, as it has the ability to adapt to new joint distributions, whereas FedEM only adapts to conditional distributions. Our approach and FedEM both surpass FedAvg+ as it is unable to adapt to new data distributions, leading to subpar performance when there is a change in the distribution. Our approach's ability to model the joint distribution with a mixture model allows for easy generalization to unseen client data, making it a practical and effective solution in cases of client drift. More results are included in Appendix \ref{app:unseen}.

\begin{table}[H]\vspace{-0.6cm}
\setlength{\tabcolsep}{1pt}
\centering
  \caption{Average test accuracy of new clients unseen at training.}\label{tab:unseen}
\vspace{-2mm}

  \begin{center}
\scalebox{0.8}{
  \begin{tabular}{|c|c |c |c|c|c|}
  \hline
 \textbf{Model}  & FedAvg  & FedAvg+   & FedEM & \ourmethod\ \\\hline
  \textbf{FEMNIST} &  74.50  & 51.00  &  72.00  & \textbf{78.51} \\\hline
  \textbf{CIFAR10} &  44.51  &  32.25 &   47.51 & \textbf{50.25} \\\hline
  \textbf{CIFAR100} &  11.50  &  7.75  &  16.50  & \textbf{21.25} \\\hline
\end{tabular}
}
\end{center}

\normalsize
\vspace{-0.8cm}
\end{table}
\section{Additional Related Work}
\vspace{-0.1in}
\label{sec:related}
There has been significant advancement in the creation of new techniques to address various FL difficulties in recent years~\cite{Wang2020Federated,kairouz2021advances,li2020federated,FederatedYaodong}. Research in this field focuses on how to do model aggregation, how to achieve personalization ~\cite{achituve2021personalized,NIPSHuili}, how to attack/defense the federated learning system~\cite{MaximilianGradient}, and efficiency aspects including communication efficiency~\cite{HierarchicalLiu, FederatedMohammad, CommunicationOsama,hou2022fedchain,hyeon-woo2022fedpara}, hardware efficiency~\cite{HAFLO} and algorithm efficiency~\cite{ICLR22Diverse,xu2022acceleration}. In this section, we focus on reviewing two groups of works: personalized federated learning and federated uncertainty quantification.

 
\vspace{-0.1in}
\subsection{Personalized Federated Learning}
\vspace{-0.05in}
However, in real settings, there always exists statistical heterogeneity across clients~\cite{kairouz2021advances,li2020federated,sattler2019robust}. There are many efforts on extending the FL methods for heterogeneous clients to achieve personalization~\cite{achituve2021personalized,NIPSHuili,t2020personalized,tan2022towards,fallah2020personalized,deng2020adaptive,hong2022efficient,NIPSWonyong}, adopting meta-learning, client clustering, multi-task learning, model interpolation, knowledge distillation, and lottery ticket hypothesis. For example, several works train a global model as a meta-model and then fine-tune the parameters for each client~\cite{sim2019investigation,jiang2019improving}, which still have difficulty for generalization~\cite{caldas2018leaf,marfoq2021federated}. Clients can be assigned to many clusters, and clients in the same cluster are assumed to use the same model via clustered FL techniques~\cite{ghosh2020efficient,shlezinger2020communication,sattler2020clustered}. As a result, the federated model will not be ideal because clients from various clusters would not share pertinent information. Another group of approaches uses multi-task learning to learn customized models in the FL environment~\cite{smith2017federated,vanhaesebrouck2017decentralized,caldas2018federated}, enabling more complex relationships between clients' models. They did not, however, take into account the diverse statistical diversity. The study in~\cite{marfoq2021federated} takes into account conditional client distribution but makes the assumption that their marginal distributions are stable. Our method, however, models the diversity of joint distributions among clients. For each client, some works attempt to jointly train a global model and a local model, but they may fail if some local distributions deviate significantly from the average distribution.~\cite{corinzia2019variational,deng2020adaptive}. \cite{shamsian2021personalized}  proposed to carry out personalization in federated learning via a hypernetwork. Similar to this, Dai et al. suggested using decentralized sparse training to generate PFL that is effective at communication~\cite{ICMLDacheng}. Some researchers addressed the heterogeneity by adopting knowledge distillation \cite{pmlr-v139-zhu21b,chen2021fedbe,NEURIPS2020_18df51b9}. 
\vspace{-0.1in}

\subsection{Uncertainty Quantification and OOD Detection for Personalized Federated Learning}
\vspace{-0.05in}
In the context of federated learning, when \textit{client drift} happens, i.e., the distribution of the data on different devices becomes increasingly dissimilar over time, it is desirable to detect novel clients or instances that are out-of-distribution. However, because it calls for unsupervised density estimation, this topic has not received much attention in the literature. Unsupervised federated clustering~\cite{Orchestra} or representation learning~\cite{zhuang2022divergenceaware} techniques have been described in several publications. However, these techniques cannot be used to directly estimate the joint distribution of instances, and it is difficult to perform OOD detection tasks with them. To address the issue, some researchers proposed a Bayesian approach to PFL. For example, FedPop~\cite{kotelevskii2022fedpop} is the first personalized FL approach that allows uncertainty quantification. Using an empirical Bayes prediction approach, FedPop enables personalization and on-device uncertainty measurement. FedPop, however, is unable to simulate the joint mixed distribution, which prevents it from addressing the joint distribution heterogeneity issue. Additionally, it is unable to carry out sample-wise uncertainty quantification.

\vspace{-0.15in}
\section{Conclusion}
\vspace{-0.05in}
In this paper, we address the challenge of \textit{joint distribution heterogeneity} in Personalized Federated Learning (PFL). Existing PFL methods mainly focus on modeling concept shift, which results in suboptimal performance when joint data distributions across clients diverge. These methods also fail to effectively address the problem of client drift, making it difficult to detect new samples and adapt to unseen client data. To tackle these issues, we propose a novel approach called \ourmethod, which uses Gaussian mixture models to fit the joint data distributions across FL devices. This approach effectively addresses the problem and allows for uncertainty quantification, making it easy to recognize new clients and samples. Furthermore, we present a federated Expectation-Maximization (EM) algorithm for learning model parameters, which is theoretically guaranteed to converge. The results of our extensive experiments on three benchmark FL datasets and a synthetic dataset show that our proposed method outperforms state-of-the-art baselines.

\clearpage
\newpage

\nocite{langley00}

\bibliography{ref}
\bibliographystyle{icml2023}

\newpage
\appendix

\onecolumn

\section{The Client-Server Training Algorithm.}

In this section, we detail our algorithm \ourmethod\ in Algorithm \ref{alg:FedGMM}. Specifically, At each round, clients and server are communicated as follows.
\begin{itemize}[itemsep=3pt,topsep=0pt,parsep=0pt]
    \item (1) the central server broadcasts the aggregated base models to all clients (line 2), including Gaussian parameters ($\bmu, \bSigma$) and supervised learning models ($\btheta$);
    \item (2) each client locally updates the parameter of the base models and the mixture weights (line 3-9) according to Equation~\eqref{eqn:e-step} and~\eqref{eqn:m-step};
    \item (3) the clients send the updated components $(\bmu^{(t)}_{m_1,c}, \bSigma^{(t)}_{m_1,c}), \btheta^{(t)}_{m_2,c}$ and the summed response $\gamma^{(t)}_{c}(m_1, m_2) = \sum_{i \in [N_c]} q^{(t)}_{\sbb_{c,i}}(m_1, m_2)$ back to the server (line 10);
    \item 4) the server aggregates the updates including Gaussian parameters and supervised component (line 12-17);
\end{itemize}

In Algorithm~\ref{alg:FedGMM-GMM}, we also provide a pure unsupervised federated (client-server) GMM algorithm. We will prove its convergence property of it in the next section. The federated learning process for the Gaussian mixture is a novel federated unsupervised learning approach, which may be of independent interest.

\label{appendix:alg}
\begin{algorithm}[ht!] 
\caption{Algorithm of FedGMM}
\begin{algorithmic}[1] \label{alg:FedGMM}
    \FOR {$t = 1,2, \dots$}
    \STATE {server broadcasts $\{\bmu^{(t-1)}_m, \bSigma^{(t-1)}_m\}_{m \in [M_1]}, \{\btheta^{(t-1)}_m \}_{m \in [M_2]}$ to all clients}
        \FOR {client $c \in [C]$}
            \FOR {component $m_1 \in [M_1], m_2 \in [M_2]$}
                \FOR {sample $\sbb_{c,i} = (\xb_{c,i}, \yb_{c,i})$, $i \in [N_c]$}
                    \STATE 
                    Set $q^{(t)}_{\sbb_{c,i}}(m_1, m_2) \propto 
                    \pi^{(t-1)}_c(m_1, m_2) 
                    \cdot \cN \big(\xb_{c,i}; \bmu_{m_1}^{(t-1)}, \bSigma_{m_1}^{(t-1)} \big) 
                    \cdot \exp \big(-L_{\text{CE}}( s_{c,i}; \btheta^{(t-1)}_{m_2}) \big)
                    $ 
                \ENDFOR
                \STATE Set for all $m_1 \in [M_1], m_2 \in [M_2]:$
                    \begin{align*}
    \pi^{(t)}_c(m_1, m_2) & = \frac{1}{N_c} \sum_{i \in [N_c]} q^{(t)}_{\sbb_{c,i}}(m_1, m_2) 
    \\
    \bmu^{(t)}_{m_1,c}
    & =
    \frac{\sum_{i \in [N_c]} \sum_{m_2 \in [M_2]} q^{(t)}_{\sbb_{c,i}}(m_1, m_2)   \xb_{c,i} }{\sum_{i \in [N_c]} \sum_{m_2 \in [M_2]} q^{(t)}_{\sbb_{c,i}}(m_1, m_2) } 
    \\
    \bSigma^{(t)}_{m_1,c}
    & =
    \frac{\sum_{i \in [N_c]} \sum_{m_2 \in [M_2]} q^{(t)}_{\sbb_{c,i}}(m_1, m_2) ( \xb_{c,i}- \bmu^{(t)}_{m_1,c})( \xb_{c,i}- \bmu^{(t)}_{m_1,c})^{\top}}{\sum_{i \in [N_c]} \sum_{m_2 \in [M_2]} q^{(t)}_{\sbb_{c,i}}(m_1, m_2)} 
    \\
    \btheta^{(t)}_{m_2,c} & = 
    \arg \min_{\btheta} \sum_{i \in [N_c]} \sum_{m_1 \in [M_1]} q^{(t)}_{\sbb_{c,i}}(m_1, m_2)  L_{\text{CE}}(\xb_{c,i},\yb_{c,i}; \btheta)
\end{align*} 
            \ENDFOR
            \STATE client $c$ sends $\{\bmu^{(t)}_{m_1,c}, \bSigma^{(t)}_{m_1,c}, \gamma^{(t)}_{c}(m_1, m_2) = \sum_{i \in [N_c]} q^{(t)}_{\sbb_{c,i}}(m_1, m_2)  \}$ to the server
        \ENDFOR
        \FOR {Gaussian component $m_1 \in [M_1]$}
            \STATE server aggregates
            \begin{align*}
    \bmu^{(t)}_{m_1}
    & =
    \frac{\sum_{c \in [C]} \sum_{m_2 \in [M_2] } \gamma^{(t)}_{c}(m_1, m_2) \bmu^{(t)}_{m_1,c} }{\sum_{c \in [C]}\sum_{m_2 \in [M_2] } \gamma^{(t)}_{c}(m_1, m_2) } 
    \\
    \bSigma^{(t)}_{m_1}
    & =
    \frac{\sum_{c \in [C]} \sum_{m_2 \in [M_2] } \gamma^{(t)}_{c}(m_1, m_2) \bSigma^{(t)}_{m_1,c} }{\sum_{c \in [C]} \sum_{m_2 \in [M_2] }  \gamma^{(t)}_{c}(m_1, m_2)} 
\end{align*} 
        \ENDFOR
        \FOR {Supervised component $m_2 \in [M_2]$}
        \STATE server aggregates
        \begin{align*}
        \btheta^{(t)}_{m_2}
        & =
        \frac{\sum_{c \in [C]} \sum_{m_1 \in [M_1] } \gamma^{(t)}_{c}(m_1, m_2) \btheta^{(t)}_{m_2,c} }{\sum_{c \in [C]} \sum_{m_1 \in [M_1] } \gamma^{(t)}_{c}(m_1, m_2)} 
        \end{align*}
        \ENDFOR 
    \ENDFOR
\end{algorithmic}
\end{algorithm}

\begin{algorithm}[ht!]
\caption{Federated GMM (Unsupervised)} 
\begin{algorithmic}[1] \label{alg:FedGMM-GMM}
    \FOR {$t = 1,2, \dots$}
    \STATE {server broadcasts $\{\bmu^{(t-1)}_m, \bSigma^{(t-1)}_m\}_{m \in [M]}$ to all clients}
        \FOR {client $c \in [C]$}
            \FOR {component $m \in [M]$}
                \FOR {sample $\sbb_{c,i} = (\xb_{c,i}, \yb_{c,i})$, $i \in [N_c]$}
                    \STATE 
                    Set $q^{(t)}_{\sbb_{c,i}}(m) \propto 
                    \pi^{(t-1)}_c(m) 
                    \cdot \cN \big(\xb_{c,i}; \bmu_{m}^{(t-1)}, \bSigma_{m}^{(t-1)} \big) 
                    $ 
                \ENDFOR
                \STATE Set for all $m \in [M]:$
                    \begin{align*}
    \pi^{(t)}_c(m) & = \frac{1}{N_c} \sum_{i \in [N_c]} q^{(t)}_{\sbb_{c,i}}(m) 
    \\
    \bmu^{(t)}_{m,c}
    & =
    \frac{\sum_{i \in [N_c]} q^{(t)}_{\sbb_{c,i}}(m)   \xb_{c,i} }{\sum_{i \in [N_c]} q^{(t)}_{\sbb_{c,i}}(m) } 
    \\
    \bSigma^{(t)}_{m,c}
    & =
    \frac{\sum_{i \in [N_c]}q^{(t)}_{\sbb_{c,i}}(m) ( \xb_{c,i}- \bmu^{(t)}_{m,c})( \xb_{c,i}- \bmu^{(t)}_{m,c})^{\top}}{\sum_{i \in [N_c]} q^{(t)}_{\sbb_{c,i}}(m)}
\end{align*} 
            \ENDFOR
            \STATE client $c$ sends $\{\bmu^{(t)}_{m_1,c}, \bSigma^{(t)}_{m_1,c}, \gamma^{(t)}_{c}(m) = \sum_{i \in [N_c]} q^{(t)}_{\sbb_{c,i}}(m)  \}$ to the server
        \ENDFOR
        \FOR {Gaussian component $m \in [M]$}
            \STATE server aggregates
            \begin{align*}
    \bmu^{(t)}_{m}
    & =
    \frac{\sum_{c \in [C]} \gamma^{(t)}_{c}(m)  \bmu^{(t)}_{m,c} }{\sum_{c \in [C]} \gamma^{(t)}_{c}(m) } 
    \\
    \bSigma^{(t)}_{m}
    & =
    \frac{\sum_{c \in [C]} \gamma^{(t)}_{c}(m) \bSigma^{(t)}_{m,c} }{\sum_{c \in [C]} \gamma^{(t)}_{c}(m)} 
\end{align*} 
        \ENDFOR
    \ENDFOR
\end{algorithmic}
\end{algorithm}


\section{Proof of Theorem~\ref{thm:gmm}.}\label{sec:appproof}
In this section, we provide theoretical proof for Theorem~\ref{thm:gmm}, that indicating the log-likelihood in our proposed federated EM algorithm will finally converge to a maximum.
Before presenting the proof, we first define the surrogate function and present two lemmas regarding the monotonicity of the updates with respect to the surrogate function.

First, we lower bound the likelihood $F$ with surrogate function $G$'s as:
\begin{align*}
    F(\bmu_{1:M}, \pi_{1:C})  & = \sum_{c=1}^{C}\sum_{i=1}^{N_c}\log\bigg( \sum_{m=1}^{M} \pi^{(t)}_c(m) \cN(\xb_{c,i}; bmu^{(t)}_m, Ib) \bigg)    \notag \\
    & \ge     \sum_{c=1}^{C}    \sum_{i=1}^{N_c}    \sum_{m=1}^{M}     q_{\sbb_{c,i}}^{(t)}(m)    \Big[     \log    \big( \pi_c(m) \big) +    \log    \big( \cN(\xb_{c,i}; \bmu_m, \Ib) \big)    -    \log    \big( q_{\sbb_{c,i}}^{(t)}(m) \big)    \Big]   \notag \\
    & =    \sum_{c=1}^{C}    \underbrace{    \sum_{i=1}^{N_c}    \sum_{m=1}^{M} q_{\sbb_{c,i}}^{(t)}(m)    \bigg[     \log \big( \pi_c(m) \big) +     \frac{d \log(2\pi)}{2} -    \frac{1}{2}    \| \xb_{c,i} - \bmu_m \|_2^2 -    \log    \big( q_{\sbb_{c,i}}^{(t)}(m) \big) \bigg]}_{G^{(t)}_c(\bmu_{1:M}, \pi_c) } ,
\end{align*}
where the first inequality is due to Jensen's inequality. 
In other words, we have for any time step $t>0$
\begin{align*}
    F(\bmu_{1:M}, \pi_{1:C}) \ge G^{(t)}(\bmu_{1:M}, \pi_{1:C}) := \sum_{c=1}^{C} G^{(t)}_c(\bmu_{1:M}, \pi_c).
\end{align*} 

The inequality becomes equality when $q_{\sbb_{c,i}}^{(t)}(m) \propto \pi_c(m) \cN(\xb_{c,i}; \bmu_m, \Ib)$, that is, when the E-step is performed. 
Therefore, we have $F(\bmu_{1:M}^{(t-1)}, \pi_{1:C}^{(t-1)}) = G_c^{(t)}(\bmu_{1:M}^{(t-1)}, \pi_{1:C}^{(t-1)})$.


\begin{lemma} \label{lemma:1}
At any time step $t$, $G^{(t)}(\bmu_{1:M}^{(t)}, \pi_{1:C}^{(t-1)}) \ge G^{(t)}(\bmu_{1:M}^{(t-1)}, \pi_{1:C}^{(t-1)})$.
\end{lemma}
\begin{proof}


Notice that 
\begin{align*}
    \bmu^{(t)}_{m}
    & =
    \frac{\sum_{c \in [C]} \gamma^{(t)}_{c}(m)  \bmu^{(t)}_{m,c} }{\sum_{c \in [C]} \gamma^{(t)}_{c}(m) } 
    \\
    & =
    \frac{\sum_{c \in [C]} \sum_{i \in [N_c]} q^{(t)}_{\sbb_{c,i}}(m)  \bmu^{(t)}_{m,c} }{\sum_{c \in [C]} \sum_{i \in [N_c]} q^{(t)}_{\sbb_{c,i}}(m) }
    \\
    & =
    \frac{\sum_{c \in [C]} \sum_{i \in [N_c]} q^{(t)}_{\sbb_{c,i}}(m)  \frac{\sum_{i \in [N_c]} q^{(t)}_{\sbb_{c,i}}(m)   \xb_{c,i} }{\sum_{i \in [N_c]} q^{(t)}_{\sbb_{c,i}}(m) }  }{\sum_{c \in [C]} \sum_{i \in [N_c]} q^{(t)}_{\sbb_{c,i}}(m) } 
    \\
    & =
    \frac{\sum_{c \in [C]} \sum_{i \in [N_c]} q^{(t)}_{\sbb_{c,i}}(m)  \xb_{c,i}   }{\sum_{c \in [C]} \sum_{i \in [N_c]} q^{(t)}_{\sbb_{c,i}}(m) }, 
\end{align*}
where the first and the second equation come from the definition of $\bmu^{(t)}_{m}$ and $\bmu^{(t)}_{m,c}$, respectively.

It is easy to verify that, $\bmu^{(t)}_{m}$ is the minimizer of the objective$\sum_{c=1}^{C}
    \sum_{i=1}^{N_c}
    q_{\sbb_{c,i}}^{(t)}(m)
    \| \xb_{c,i} - \bmu \|_2^2$. 
Therefore, we have
\begin{align} 
    \sum_{c=1}^{C}
    \sum_{i=1}^{N_c}
    q_{\sbb_{c,i}}^{(t)}(m)
    \| \xb_{c,i} - \bmu^{(t)}_{m} \|_2^2
    & \le 
    \sum_{c=1}^{C}
    \sum_{i=1}^{N_c}
    q_{\sbb_{c,i}}^{(t)}(m)
    \| \xb_{c,i} - \bmu^{(t-1)}_{m} \|_2^2.
\end{align}
And further,
\begin{align*}
    G^{(t)}(\bmu^{(t)}_{1:M}, \pi^{(t-1)}_{1:C}) 
    & := 
    \sum_{c=1}^{C} \sum_{i=1}^{N_c}
    \sum_{m=1}^{M} q_{\sbb_{c,i}}^{(t)}(m)
    \bigg[ 
    \log \big( \pi^{(t-1)}_c(m) \big)
    +
    \frac{d \log(2\pi)}{2}
    -
    \frac{1}{2}
    \| \xb_{c,i} - \bmu^{(t)}_m \|_2^2
    -
    \log
    \big( q_{\sbb_{c,i}}^{(t)}(m) \big)
    \bigg]
    \\
    & \ge  
    \sum_{c=1}^{C}
    \sum_{i=1}^{N_c}
    \sum_{m=1}^{M} q_{\sbb_{c,i}}^{(t)}(m)
    \bigg[ 
    \log \big( \pi^{(t-1)}_c(m) \big)
    +
    \frac{d \log(2\pi)}{2}
    -
    \frac{1}{2}
    \| \xb_{c,i} - \bmu^{(t-1)}_{m} \|_2^2
    -
    \log
    \big( q_{\sbb_{c,i}}^{(t)}(m) \big)
    \bigg]
    \\
    & =
    G^{(t)}(\bmu^{(t-1)}_{1:M}, \pi^{(t-1)}_{1:C}).
\end{align*}
    
\end{proof}

\begin{lemma} \label{lemma:2}
At any time step $t$, $G^{(t)}(\bmu_{1:M}^{(t)}, \pi_{1:C}^{(t)}) \ge G^{(t)}(\bmu_{1:M}^{(t)}, \pi_{1:C}^{(t-1)})$.
\end{lemma}
\begin{proof}

Notice that 
\begin{align*}
    G^{(t)}_c(\bmu_{1:M}, \pi_c)
    &: = 
    \sum_{i=1}^{N_c}
    \sum_{m=1}^{M} q_{\sbb_{c,i}}^{(t)}(m)
    \Big[ 
    \log
    \big( \pi_c(m) \big)
    +
    \log
    \big( \cN(\xb_{c,i}; \bmu_m, \Ib) \big)
    -
    \log
    \big( q_{\sbb_{c,i}}^{(t)}(m) \big)
    \Big].
\end{align*}
We have for any $\pi$ and any $\bmu_{1:M}$,
\begin{align*}
    G^{(t)}_c(\bmu_{1:M}, \pi^{(t)}_{c}) - G^{(t)}_c(\bmu_{1:M}, \pi) 
    & = 
    \sum_{i=1}^{N_c}
    \sum_{m=1}^{M} 
    q_{\sbb_{c,i}}^{(t)}(m)
    \Big[ 
    \log
    \big( \pi^{(t)}_c(m) \big)
    -
    \log
    \big( \pi_c(m) \big)
    \Big]
    \\
    & =
    N_c
    \sum_{m=1}^{M} 
    \frac{1}{N_c}
    \sum_{i=1}^{N_c}
    q_{\sbb_{c,i}}^{(t)}(m)
    \Big[ 
    \log
    \big( \pi^{(t)}_c(m) \big)
    -
    \log
    \big( \pi_c(m) \big)
    \Big]
    \\
    & = 
    N_c
    \sum_{m=1}^{M} 
    \pi^{(t)}_c(m)
    \Big[ 
    \log
    \big( \pi^{(t)}_c(m) \big)
    -
    \log
    \big( \pi_c(m) \big)
    \Big]
    \\
    & = 
    N_c \cdot \mathrm{KL}(\pi^{(t)}_c \| \pi_c) \ge 0,
\end{align*}
where the third equation comes from the definition of $\pi^{(t)}_c$, and the last equation comes from the definition of the KL-divergence.

Therefore, we have
\begin{align*}
    G^{(t)}(\bmu_{1:M}^{(t)}, \pi_{1:C}^{(t)}) 
    & = 
    \sum_{c=1}^{C} G^{(t)}_c(\bmu_{1:M}^{(t)}, \pi_{c}^{(t)}) 
    \\
    & \ge 
    \sum_{c=1}^{C} G^{(t)}_c(\bmu_{1:M}^{(t)}, \pi_{c}^{(t-1)}) 
    \\
    & =
    G^{(t)}(\bmu_{1:M}^{(t)}, \pi_{1:C}^{(t-1)}).
\end{align*}
\end{proof}

\begin{proof}[Proof of Theorem~\ref{thm:gmm}]
By Lemma~\ref{lemma:1} and Lemma~\ref{lemma:2}, we have for any $t>0$,
\begin{align*}
    G^{(t)}(\bmu_{1:M}^{(t)}, \pi_{1:C}^{(t)}) \ge G^{(t)}(\bmu_{1:M}^{(t-1)}, \pi_{1:C}^{(t-1)}),
\end{align*}
which further gives:
\begin{align*}
    F(\bmu_{1:M}^{(t)}, \pi_{1:C}^{(t)}) \ge G^{(t)}(\bmu_{1:M}^{(t)}, \pi_{1:C}^{(t)}) \ge G^{(t)}(\bmu_{1:M}^{(t-1)}, \pi_{1:C}^{(t-1)})
    =
    F(\bmu_{1:M}^{(t-1)}, \pi_{1:C}^{(t-1)}).
\end{align*}
Here, the first inequality holds because $G^{(t)}$ is a surrogate that always satisfies $F(\cdot) \ge G^{(t)}(\cdot)$; the last equation holds as we discussed at the beginning of this section.

This actually shows that $F(\bmu_{1:M}^{(t)}, \pi_{1:C}^{(t)}) $ is monotonically increasing, and since $F(\bmu_{1:M}^{(t)}, \pi_{1:C}^{(t)}) $ is upper bounded by some constant $F^*$, it is easy to show 
\begin{align*}
    \frac{1}{T}
    \sum_{t=2}^{T}
    |F(\bmu_{1:M}^{(t)}, \pi_{1:C}^{(t)})  - F(\bmu_{1:M}^{(t-1)}, \pi_{1:C}^{(t-1)}) |
    & \le 
    \frac{1}{T} (F^* - F(\bmu_{1:M}^{(1)}, \pi_{1:C}^{(1)}) ) = O(T^{-1}).
\end{align*}

\end{proof}

\section{Appendix for  Experiments.}
\label{sec:appExp}
\subsection{Details of Training Configuration}
\label{sec:appImp}

{\bf Hardware and Implementations.}
In this paper, we implemented our method on a Linux machine with 8 NVIDIA A100 GPUs, each with 80GB of memory. The software environment is CUDA 11.6 and Driver Version 520.61.05. We used Python 3.9.13 and Pytorch 1.12.1 ~\cite{paszke2019pytorch} to construct our project. 

{\bf Hyperparameters, Architecture, and Dataset Split.}
In our experiments, we use grid search to obtain the best performance. We provide all of the hyperparameters as well as their configurations in the following:

\begin{itemize}
    \item Optimizer: SGD is chosen as the local solver, as in~\cite{marfoq2021federated}. For each method, we follow~\cite{marfoq2021federated} to tune the learning rate via grid search in the range $\{10^{-0.5},10^{-1},10^{-1.5},10^{-2},10^{-2.5},10^{-3}\}$ to obtain the best performances. For our proposed \ourmethod, the learning rate is set to $0.01$ on CIFA-R10, $0.001$ on CIFAR-100 and FEMNIST. 
    
    \item Number of Components: $M_1$ and $M_2$ of \ourmethod\ are tuned via grid search. For our method $M_1$=3 and $M_2$=3. The setting is consistent with the setting of FedEM.
    
    \item Epochs and Batch Size: The total communication round is set to 200, and the batch size is set to 128.
    
    \item Supervised Learning Model Architecture: For fairness, for all baseline methods, including Local, FedAvg \citep{mcmahan2017communication}, FedProx \citep{li2020federated}, FedAvg+ \citep{jiang2019improving} and Clustered FL \citep{sattler2020clustered}, pFedMe~\cite{t2020personalized} and FedEM~\cite{marfoq2021federated}, the supervised backbone is the same as ours. Following \citep{marfoq2021federated}, we apply MobileNet-v2 as the supervised encoder backbone for CIFAR-10 and CIFAR-100 datasets. For FEMNIST, we use a 2-layer CNN + 2-layer FFN as the encoder, that is two convolutional layers (with $3 \times 3$ kernels), max pooling, and dropout, followed by a 128 unit dense layer as in~\cite{reddi2020adaptive}. We use Torchvision~\cite{marcel2010torchvision} to implement the MobileNet-v2.
    
    \item Dataset Split: For training, we sub-sampled 15\% from FEMNIST datasets. Detailed dataset partitioning can be found in~\cite{marfoq2021federated}. The performance of our method is evaluated on the local test data on each client and we report the average accuracy of all clients. 
\end{itemize}

\subsection{Convergence Plots}
\begin{figure*}%
    \centering
    \subfloat[\centering CIFAR-10]{{\includegraphics[width=0.333\linewidth]{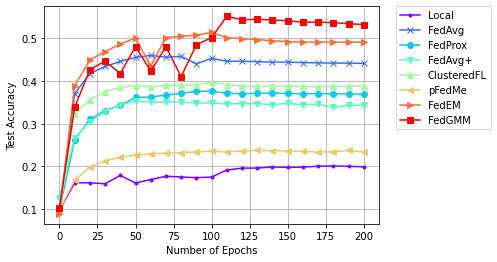} }}%
    \hspace{-2.5mm}
    \subfloat[\centering CIFAR-100]{{\includegraphics[width=0.333\linewidth]{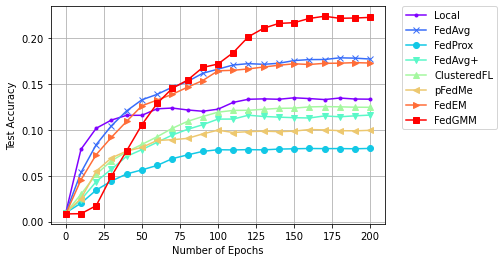} }}%
    \hspace{-2.5mm}
    \subfloat[\centering FEMNIST ]{{\includegraphics[width=0.333\linewidth]{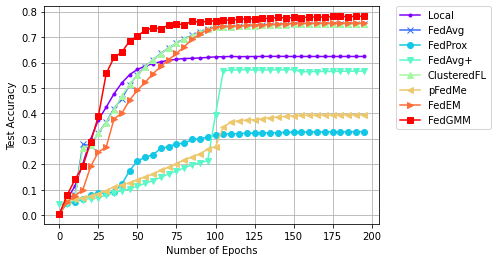} }}%
    \caption{Test accuracy on different datasets \textit{w.r.t.} training epochs.}
    \label{fig:ACCcurve}%
\end{figure*}

Figure \ref{fig:ACCcurve} shows the evolution of average test accuracy
overtime for each experiment shown in Table \ref{tab:result2}.
As shown in the table and the figure, \ourmethod\ outperforms all the baselines. This is a result of its ability to construct personalized models based on the joint data distribution, effectively capturing the heterogeneity of each sample across different clients.

\subsection{More Results on OOD Detection}
\label{app:ood}
To evaluate the OOD detection performance of \ourmethod\, we first create a federated setting of MNIST by distributing samples with the same label across the clients according to a symmetric Dirichlet distribution with parameter 0.4, as in~\cite{marfoq2021federated}. Then the overall data are equally partitioned into two sets before being further dispatched to clients. The first set of data remains unchanged, and the second set of data is further equally partitioned into two subsets: 1) In the first subset of data, we simulate heterogeneity of $\PP_c(\xb)$ by transforming sampled images with 90-degree rotation, horizontal flip, and inverse~\cite{journalsSurvey} (such transformations are denoted by $T(\cdot)$); 2) In the second subset of data, we simulate heterogeneity of $\PP_c(\yb | \xb)$ by altering labels of sampled images to a randomly generated permutation (denoted by $P_A$). 

During the evaluation stage, we examine whether a model can detect a testing sample is known or novel by the following steps:
1) we create two identical sets of test samples drawn from the same distribution of training data. The first set of test data remains unchanged. For the second set of test data, we simulate a different set of heterogeneity of $\PP_c(\xb)$ by transforming sampled images with a scale factor of 0.5, 90-degree rotation, and horizontal flip~\cite{journalsSurvey}. 
2) we labeled the first set of data as in-domain data and the second set of data as out-of-domain data. 

Similar to \citep{liu2020energy}, in Figure \ref{fig:oodfre}, we visualized the normalized likelihood histogram of known and novel samples for \ourmethod, FedEM, and FedAvg. The figures indicate the likelihoods of \ourmethod\ are more distinguishable for known and novel samples than the baselines. 

To further demonstrate the effectiveness of \ourmethod,  we visualized the frequency of samples {\em w.r.t.} the normalized likelihood against $\PP(\bx)$ and $\PP(\by|\bx)$. For perturbing $\PP(\bx)$, we only simulated a different set of heterogeneity of $\PP_c(\xb)$ by transforming sampled images with a scale factor of 0.8, and 90-degree rotation~\cite{journalsSurvey}. For perturbing $\PP(\by|\bx)$, we only altered the labels of sampled images to a randomly generated permutation. The figures indicate the joint likelihood of \ourmethod\ are more distinguishable against the changes of $\PP(\bx)$ but slightly less distinguishable against the changes of $\PP(\by|\bx)$. 

\begin{figure*}%
\vspace{-0.6cm}
    \centering
    \subfloat[\centering FedAvg]{{\includegraphics[width=0.33\linewidth]{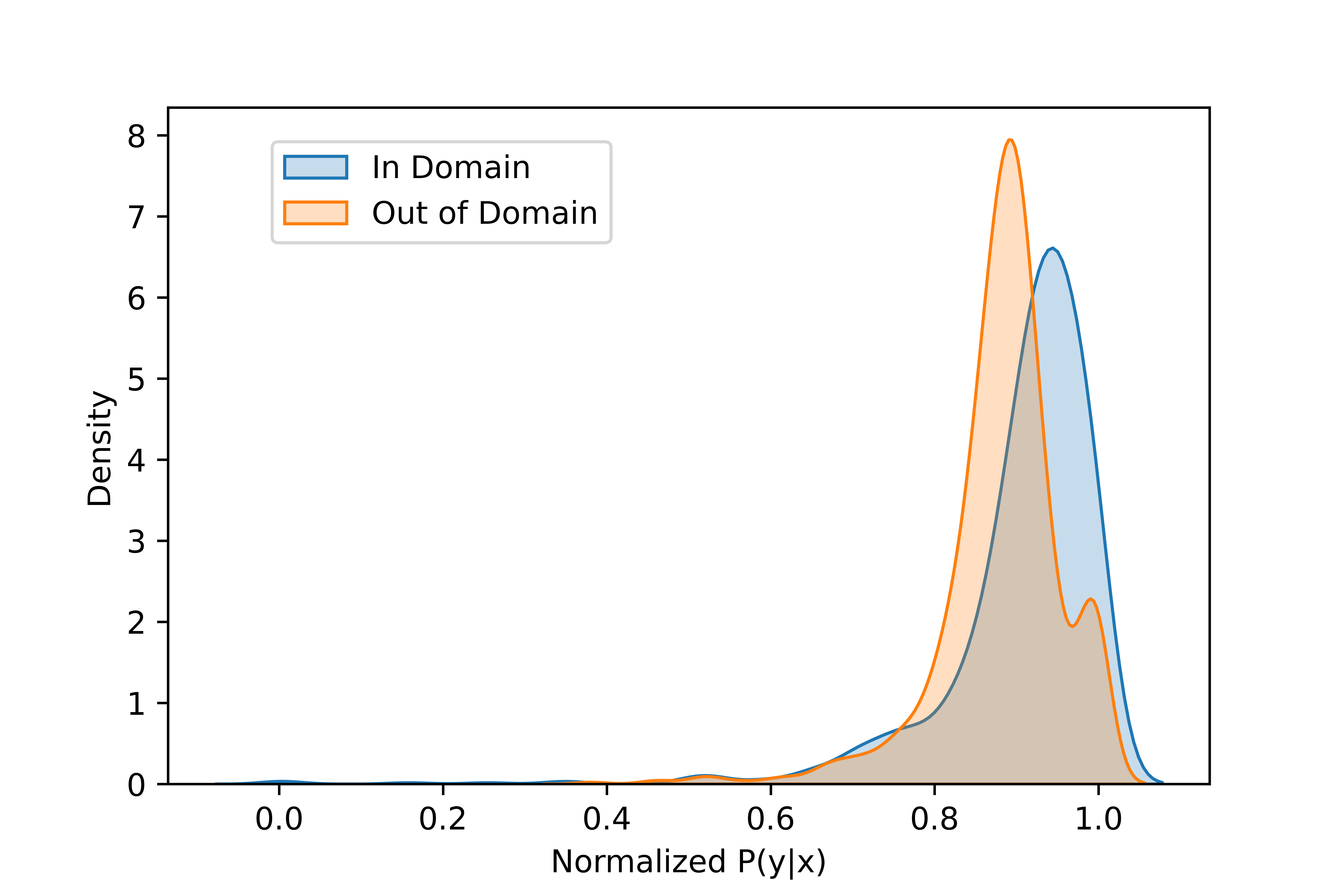} }}%
    \hspace{-2.5mm}
    \subfloat[\centering FedEM]{{\includegraphics[width=0.33\linewidth]{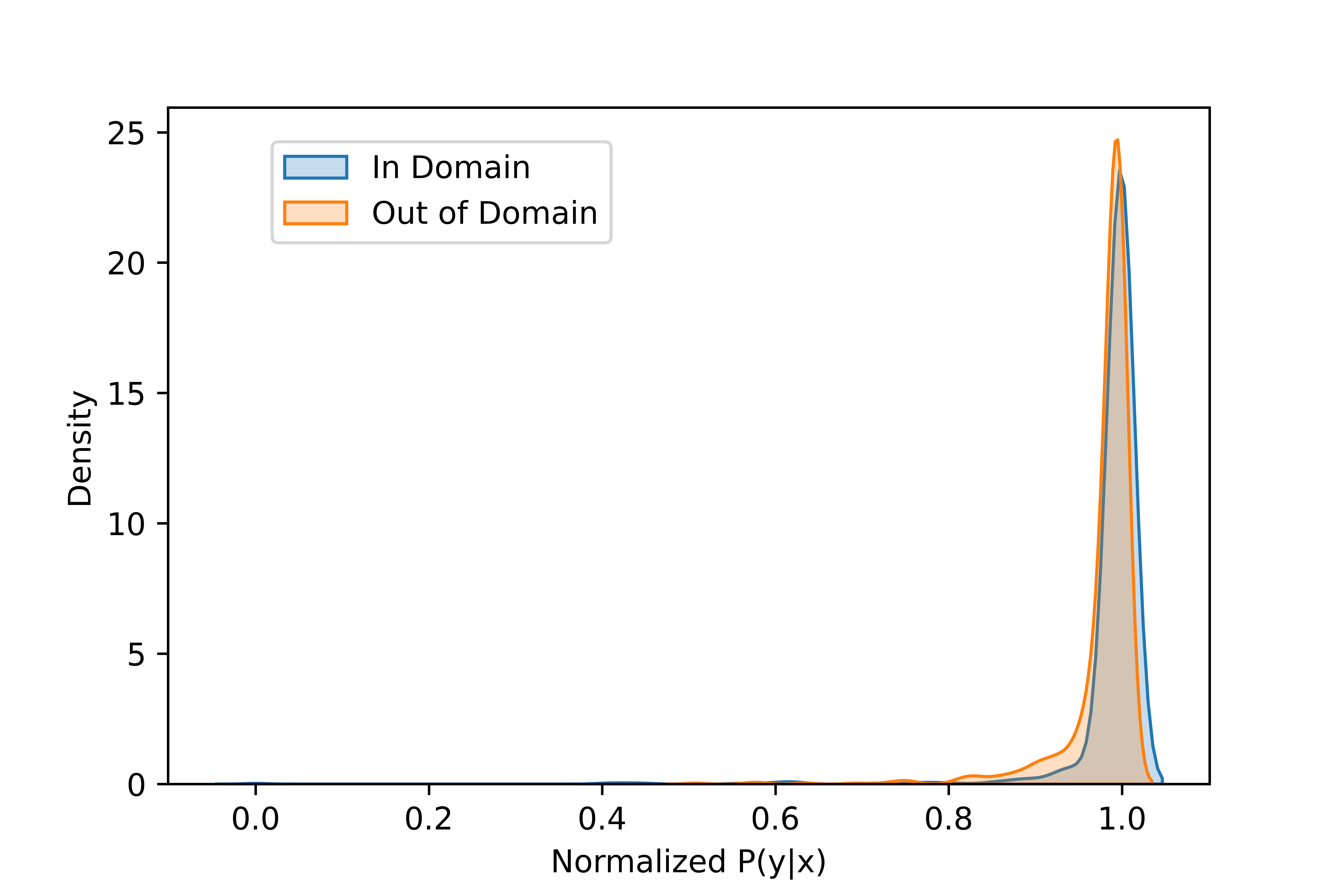} }}%
    \hspace{-2.5mm}
    \subfloat[\centering \ourmethod\ ]{{\includegraphics[width=0.33\linewidth]{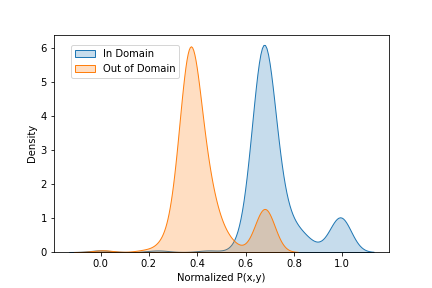} }}%
    \caption{The frequency of samples {\em w.r.t.} the normalized likelihood for (a) FedAvg (b) FedEM and (c) \ourmethod.}
    \label{fig:oodfre}%
\end{figure*}

\begin{figure*}%
\vspace{-0.6cm}
    \centering
    \subfloat[Perturbed $\PP(\bx)$]{{\includegraphics[width=0.33\linewidth]{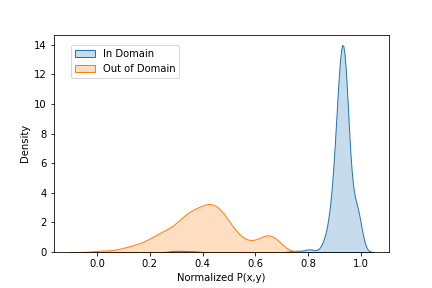} }}%
    \hspace{-2.5mm}
    \subfloat[Perturbed $\PP(\by|\bx)$ ]{{\includegraphics[width=0.33\linewidth]{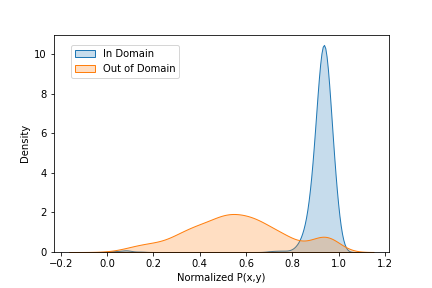} }}%
    \caption{The frequency of samples {\em w.r.t.} the normalized likelihood for \ourmethod\ on perturbed $\PP(\bx)$ and $\PP(\by|\bx)$.}
    \label{fig:oodfre}%
\end{figure*}

\begin{figure}[H] 
\vspace{-0.14in}
 \center{\includegraphics[width=7cm]  {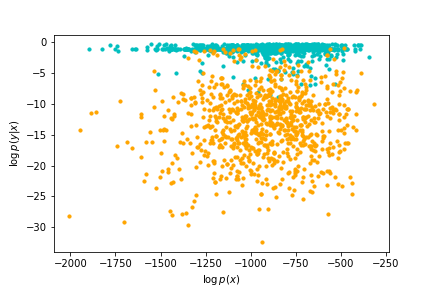}} 
 \caption{\label{fig:oodplot} $\log\PP(\bx)$ vs $\log \PP(\by|\bx)$ \emph{w.r.t.} change of $\PP(\by|\bx)$.} \label{fig:ood2}
 \end{figure}

\subsection{More Results on Effectiveness on Non-Gaussian Distribution Data}
\label{app:nonGaussian}
Besides, to see how the simulation results would change if we deviate from Gaussian assumptions, we conducted the following synthetic experiments. We use two settings to conduct the comparison. Setting 1 considers non-Gaussian input distribution. Setting 2 is also a synthetic setting, where some of the clients completely differ from others. Specifically,  Setting 1 is the same as our Gaussian synthetic setting, but the data-generating distribution is different. Here, we adopt two different distributions, i.e., Laplace and Beta distributions. Other distributes would be similar. First, we generate 3 $d$-dimensional ($d=32$) components based on the selected distribution type. Each component determined either by the mean vector $\mu$ for Laplace distribution or the vectors $\alpha$ and $\beta$ for Beta distribution. Then, we generate data from these components using multivariate distribution. We use Dirichlet distribution to distribute data to each client. Totally, we have 30 clients. For Setting 2, some clients sampled data from Gaussian, the others from a different distribution (i.e., Laplace or Beta distribution). Similarly, we also use
30 clients for simulation. The first 20 clients' data are sampled from Gaussian, and the data of the last 10 clients are sampled from selected distribution, i.e., Laplace or Beta distribution. We use Dirichlet distribution to distribute data to each client.
The results are summarized in Table. \ref{tab:nonguassian}. From the table, we can observe that under both settings, our method can still perform well since our model considers the cluster and mixture structure of the data distribution.
\begin{table}[!ht]
    \centering
    \caption{\label{tab:nonguassian} Effectiveness on Non-Gaussian Distribution Data}
    \begin{tabular}{|c|c|c|c|c|}
    \hline
         & \multicolumn{2}{c|}{\textbf{Setting 1}} & \multicolumn{2}{c|}{\textbf{Setting 2}}\\ \hline
         & Beta & Laplace & Beta/partial & Laplace/partial \\ \hline
        \textbf{FedGMM(Ours)} & \textbf{72.12} &\textbf{ 89.06} & \textbf{80.54} & \textbf{84.79 }\\ \hline
        \textbf{FedEM} & 71.77 & 83.94 & 74.22 & 81.79 \\ \hline
        \textbf{FedAVG} & 56.24 & 82.45 & 56.13 & 70.15 \\ \hline
        \textbf{FedAVG+Local} & 56.6 & 82.53 & 57.7 & 70.36 \\ \hline
        \textbf{fedProx} & 55.64 & 75.64 & 55.9 & 71.16 \\ \hline
        \textbf{ClusteredFL} & 56.23 & 82.45 & 56.1 & 70.14 \\ \hline
        \textbf{Local} & 58.46 & 83.68 & 67.18 & 74.69 \\ \hline
    \end{tabular}
\end{table}

\subsection{More Results on Adaptation to Unseen Clients}
\label{app:unseen}
\begin{figure}[H] 
\vspace{-0.14in}
\centering
    \subfloat[\centering CIFAR-10]{{\includegraphics[width=0.33\linewidth]{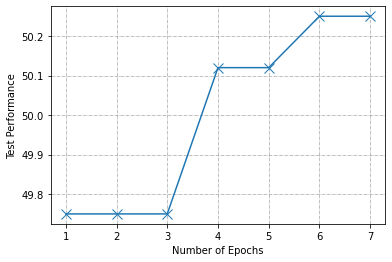} }}%
    \hspace{-2.5mm}
    \subfloat[\centering CIFAR-100]{{\includegraphics[width=0.33\linewidth]{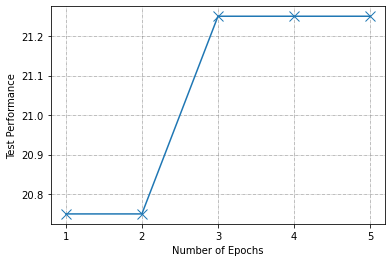} }}%
    \hspace{-2.5mm}
    \subfloat[\centering FEMNIST ]{{\includegraphics[width=0.33\linewidth]{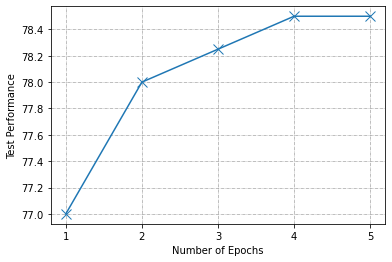} }}
 \caption{\label{fig:label_perm} Performance of FedGMM adapting to unseen clients (CIFAR-10, CIFAR-100, and FEMNIST) \textit{w.r.t.} number of epochs.}\label{fig:unseenCurve}
 \end{figure}

As discussed, FedGMM is flexible, enabling
easy inference of new clients who did not participate in
the training phase. The adaptation to unseen clients is accomplished by learning their
personalized mixture weights. Such generalization only incurs minimal computational cost. We plot the accuracy with respect to the adaptation of $\pi$ in Figure \ref{fig:unseenCurve} on different datasets, from which we can see the adaptation only needs a small computational overhead.

\subsection{Parameter Sensitivity}
\label{sec:appPrameter}
We also analyzed the hyperparameters of \ourmethod\ in this section. Basically, \ourmethod\ only has two hyper-parameters, i.e., $M_1$ and $M_2$. Different choices of the number of mixture components do not significantly impact the model's classification performance. However, the clustering quality may vary depending on the number of components used. We present the accuracy with respect to the number of GMM cluster components and supervised learning model components in Figure \ref{fig:label_perm}. The figure shows that our algorithm is not very sensitive to hyperparameters and that selecting a component number close to the ground-truth component number of the distribution can improve the clustering quality and boost the classification performance. In our setting, we have two ground-truth clusters, and labeling functions, thus the setting of $M_1$=2 and $M_2$=2 gets the best performance.

\begin{figure}[H] 
\vspace{-0.14in}
 \center{\includegraphics[width=0.5\linewidth]  {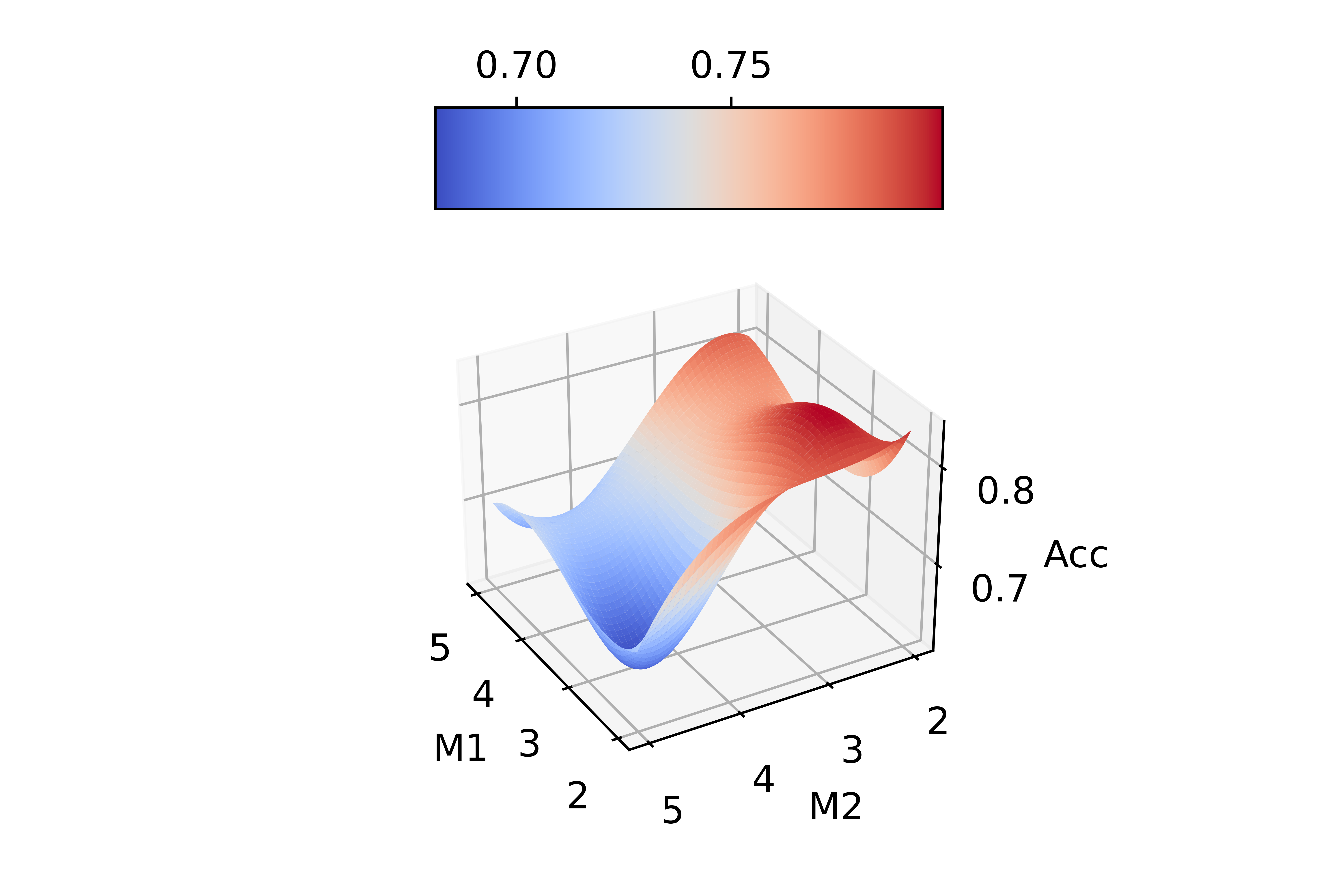}} 
 \caption{\label{fig:label_perm} Parameter sensitivity analysis with respect to the number of GMM clusters, number of classifiers, and performance.}
 \end{figure}


\end{document}